\documentclass{article}

\usepackage[preprint]{neurips_2026}

\usepackage[utf8]{inputenc} %
\usepackage[T1]{fontenc}    %
\usepackage{hyperref}       %
\usepackage{url}            %
\usepackage{booktabs}       %
\usepackage{amsfonts}       %
\usepackage{nicefrac}       %
\usepackage{microtype}      %
\usepackage{xcolor}         %
\usepackage{todonotes}  
\usepackage{natbib}
\usepackage{xspace}
\usepackage{tikz}
\usepackage{soul}
\usepackage{amsmath}
\usepackage{bm}
\usepackage[inline,shortlabels]{enumitem}
\usepackage{wrapfig}
\usepackage{multirow}
\usepackage{amssymb}
\usepackage{subcaption}
\usepackage{algorithm}
\usepackage{algpseudocode}
\usepackage{colortbl}
\usepackage{array}
\usepackage{pifont}

\usepackage{graphicx}   %
\usepackage{rotating}   %

\setuptodonotes{inline}

\def\timee{\textsc{TimEE}\xspace}
\def\halffig{0.4\textwidth}

\definecolor{orange-synth}{HTML}{FF7F0E}
\definecolor{blue-synth}{HTML}{6C8EBF}
\definecolor{grey-synth}{HTML}{B7B7B7}
\title{\timee: End-to-end Time Series Classification\\ via In-Context Learning}

\newcommand{\equal}{\textsuperscript{\rm *}}
\newcommand{\equaltext}{\equal\ Equal contribution}

\newcommand{\ufr}{\textsuperscript{\rm 1}}
\newcommand{\ufrc}{\textsuperscript{\rm 1,}}
\newcommand{\ufrtext}{\ufr\ University of Freiburg}

\newcommand{\eliza}{\textsuperscript{\rm 2}}
\newcommand{\elizac}{\textsuperscript{\rm 2,}}
\newcommand{\elizatext}{\eliza\ Zuse School ELIZA Darmstadt}

\newcommand{\prior}{\textsuperscript{\rm 3}}
\newcommand{\priorc}{\textsuperscript{\rm 3,}}
\newcommand{\priortext}{\prior\ Prior Labs}

\newcommand{\ellis}{\textsuperscript{\rm 4}}
\newcommand{\ellisc}{\textsuperscript{\rm 4,}}
\newcommand{\ellistext}{\ellis\ ELLIS Institute Tübingen}

\author{
Jaris Küken \ufrc\elizac\equal \quad Shi Bin Hoo \ufrc\equal \quad Martin Mráz \ufr \\ \textbf{Frank Hutter} \priorc\ellisc\ufr \quad \textbf{Lennart Purucker} \priorc\ufr \\ 
\ufrtext, \elizatext, \\ \priortext, \ellistext \\
\equaltext \\
\texttt{\{kuekenj,hoos\}@cs.uni-freiburg.de}
}

\begin{document}

\maketitle

\begin{abstract}
    Time series classification (TSC) is dominated by a two-stage paradigm: train a feature encoder---either from scratch on the target dataset or via pretraining on large corpora---and then fit a task-specific classifier on top.
    While effective, this decoupling optimizes representation learning independently of the classification objective, requires per-dataset training, and prevents the model from exploiting label information during inference.
    We introduce \timee \footnote{\timee pronounced as "Timmy"}, a 4.5M-parameter foundation model for end-to-end TSC via in-context learning. 
    Given a labeled support set and a query time series, \timee directly outputs a predicted class distribution in a single forward pass with no per-dataset training required.
    Following the prior-data fitted network (PFN) framework, \timee is meta-trained exclusively on synthetic TSC tasks, where each task contains time series with distinct class identities arising from structured distributional shifts in the generative process.
    Despite seeing no real time series during pre-training, \timee ranks first in ROC AUC (and third on accuracy) on the UCR benchmark among all compared methods, which include both foundation models and supervised deep learning baselines. 
    To our knowledge, \timee is the first purely synthetic-pretrained model to reach state-of-the-art performance on the UCR benchmark.
    These results establish end-to-end ICL with synthetic priors as a compelling, largely unexplored direction for TSC, with scaling, prior design, and richer generation mechanisms as natural avenues for improvement.
    Code is publicly available at \url{https://github.com/automl/timee}.

\end{abstract}

\section{Introduction}

The dominant approach to time series classification (TSC) decomposes the problem into two stages: first, a feature encoder maps each time series to a fixed-dimensional representation; second, a task-specific classifier is trained on these representations to produce predictions. Despite the surface-level differences of existing methods, this two-stage design is universal. Classical methods such as MiniRocket~\citep{dempster2021minirocket} and Hydra~\citep{dempster2023hydra}, per-dataset deep learning methods such as TS2Vec~\citep{yue2022ts2vecuniversalrepresentationtime}, and recent foundation models such as 
Chronos-2~\citep{ansari_chronos-2_2025}, TiRex~\citep{auer2025tirexzeroshotforecastinglong}, and
MantisV2~\citep{feofanov2026mantisv2closingzeroshotgap} all instantiate it.
\\
We argue that this is not merely an implementation convention, but a structural bottleneck.
The encoder is optimized independently of the classification objective, and a new classifier must be fitted for every new dataset.
Most critically, at inference time, the encoder cannot directly access the labeled training samples and therefore cannot produce representations tailored to the dataset's class structure.

In-context learning (ICL) eliminates all three of these downsides simultaneously.
Rather than fitting per-dataset weights, an ICL model observes the labeled training set at inference and directly outputs a predicted class distribution in a single forward pass.
No weight updates are required.
This paradigm has proven transformative for tabular classification: TabPFN~\citep{hollmann-iclr23a, hollmann-nature25a, grinsztajn-arXiv26a} and TabICL~\citep{qu2026tabiclv2betterfasterscalable, qu2025tabicltabularfoundationmodel} achieve state-of-the-art performance by meta-training over synthetic labeled tasks drawn from a structured prior, with no per-dataset training at test time.

Extending this to TSC, however, requires solving a non-trivial data generation problem. The tabular ICL literature has largely resolved synthetic pre-training through structured Bayesian priors over causal graphs. 
Existing synthetic generation approaches for time series either produce unlabeled corpora~\citep{xie_cauker_2025} or assign class membership arbitrarily~\citep{yeh2025tictsyntheticallypretrainedfoundation}.
Neither approach provides the structured, label-meaningful datasets that ICL meta-training requires to learn a classification prior. To address this problem, we introduce a novel synthetic data generation pipeline. 

We introduce \timee, a 4.5M-parameter model for end-to-end TSC via ICL.
\timee is meta-trained exclusively on synthetic, labeled time series classification tasks and requires no per-dataset training at test time.
Despite seeing no real time series during pre-training, \timee achieves the lowest mean rank on ROC AUC across all 128 UCR benchmark datasets, outperforming all compared methods, including both foundation models and supervised baselines.
\timee lies on the Pareto front of inference speed versus predictive performance.

\textbf{Our contributions are:}
\begin{enumerate*}[font=\bfseries,label=(\Roman*)]
    \item a new end-to-end ICL paradigm for TSC requiring no per-dataset training;
    \item a transformer architecture that jointly models temporal and cross-series structure for in-context classification;
    \item a VARX-based synthetic prior that generates labeled TSC tasks grounded in the data-generating process;
    \item state-of-the-art results on the UCR benchmark with strong probabilistic calibration and competitive inference speed.
\end{enumerate*}

\begin{figure}
    \centering
    \includegraphics[width=\linewidth]{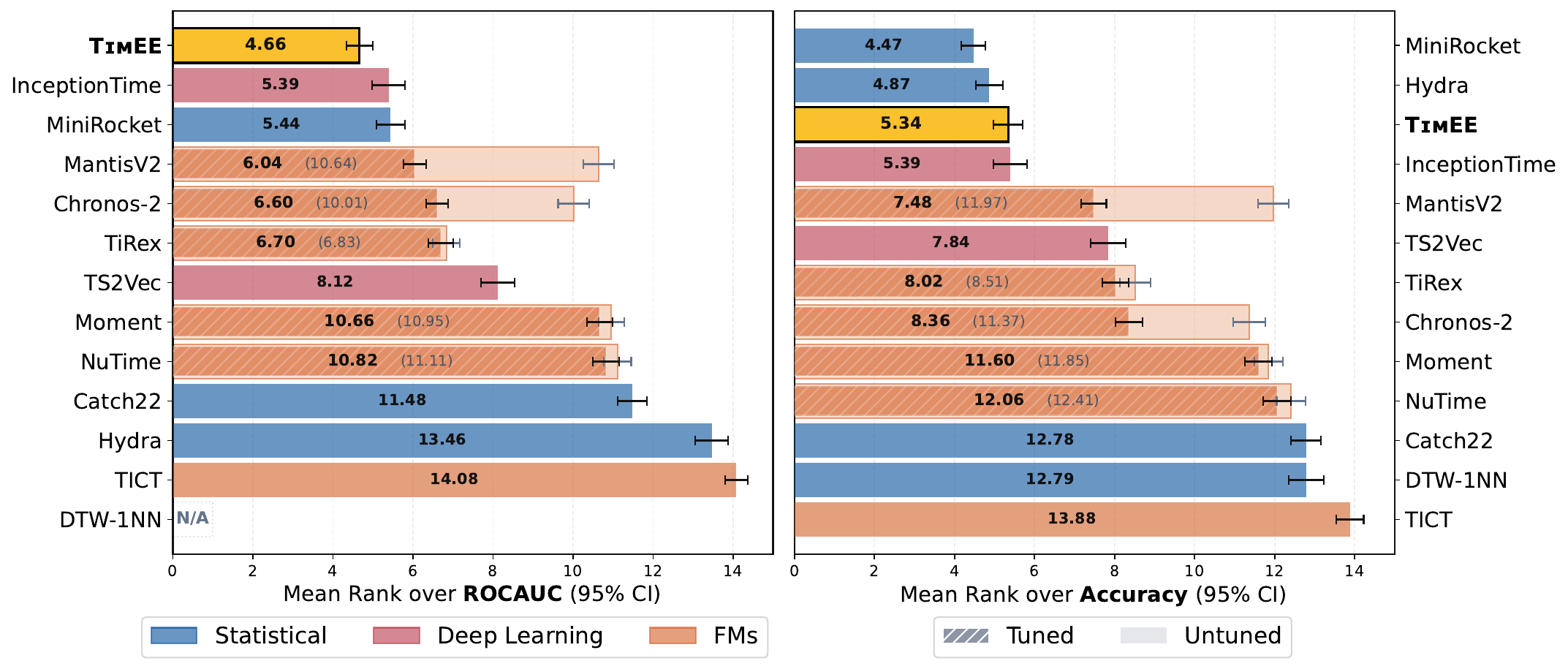}
    \caption{
    We compare mean rank across all 128 UCR benchmark datasets.
    \timee achieves the lowest mean rank on ROC AUC (left), outperforming all compared methods including task-specific and foundation model baselines, while ranking third on accuracy (right).
    DTW-1NN is excluded from the ROC AUC comparison as it does not produce calibrated probability scores.
}
    \label{fig:main-res}
\end{figure}

\section{Background} \label{sec:background}
We introduce the time series classification problem and review key concepts and related work. %

\subsection{Time Series Classification}
Consider a time series classification problem with labeled training data $\mathcal{D}_{train} = {(\mathbf{x}_i, y_i)}_{i=1}^n$, where $\mathbf{x}_i \in     
\mathbb{R}^{T \times d}$ is a $d$-dimensional time series of length $T$ and $y_i \in \mathcal{Y}$ is the corresponding class label. The goal is to estimate the predictive distribution $p(\cdot \mid \mathbf{x}_{test},
\mathcal{D}_{train})$ for a held-out test sample $\mathbf{x}_{test}$.

\subsection{The Two-Stage Paradigm} \label{sec:two-stage-paradigm}
The dominant approach to TSC decomposes the problem into two stages. First, a feature encoder $F_\theta: \mathbb{R}^{T \times d} \rightarrow \mathbb{R}^m$ maps each time series to a fixed-dimensional representation.
Second, a task-specific classifier $G_\phi$ is trained on these representations to produce predictions $\hat{y} = G_\phi(F_\theta(x))$.
We describe three general classes of methods below.
A detailed comparison is provided in Appendix Table~\ref{tab:two-stage}.

\textbf{Classical and Time Series Features Engineering Methods} define $F_\theta(x)$ without gradient-based learning, either through hand-crafted time series features or random convolutional projections.
DTW-1NN uses dynamic time warping distance, while Catch22~\citep{lubba2019catch22canonicaltimeseriescharacteristics} extracts a curated set of 22 time series features.
Rocket~\citep{dempster2019rocket} and MiniRocket~\citep{dempster2021minirocket} transforms each series using a large bank of random convolutional kernels, aggregating features via global pooling.
Hydra~\citep{dempster2023hydra} extends this by organizes kernels into competing groups to bridge random projections and dictionary-based pattern counting.
Despite requiring no representation learning, these methods remain surprisingly competitive on standard benchmarks \cite{chen_ucr_2015}.

\textbf{Per-Dataset Deep Learning Methods} learns $F_{\theta}$ on each target dataset.
InceptionTime~\citep{Ismail_Fawaz_2020} trains a convolutional encoder jointly with $G_\phi$.
TS2Vec~\citep{yue2022ts2vecuniversalrepresentationtime} instead learns representations via hierarchical contrastive objectives over time and instance dimensions on the unlabeled target dataset, before fitting a downstream classifier head.
These methods can achieve strong per-dataset performance but require training from scratch on every new dataset with no cross-task transfer.

\textbf{Time Series Foundation Models} amortize representation learning through large-scale pre-training.
TSC-focused models, namely NuTime~\citep{lin2024nutimenumericallymultiscaledembedding}, MOMENT~\citep{goswami_moment_2024}, Mantis~\citep{feofanov_mantis_2025}, and MantisV2~\citep{feofanov2026mantisv2closingzeroshotgap}, pre-train encoders via contrastive or reconstruction objectives on large corpora of real and/or synthetic time series data, producing general-purpose embeddings that can be passed to lightweight downstream classifiers.
As a complementary line of work, \citet{auer2025pretrainedforecastingmodelsstrong} studied transferring forecasting foundation models (notably Chronos-2~\citep{ansari_chronos-2_2025} and TiRex~\citep{auer2025tirexzeroshotforecastinglong}) to classification using the extracted representations.
In practice, however, competitive performance across all these models requires non-trivial layer selection: intermediate representations systematically outperform final-layer embeddings~\citep{auer2025pretrainedforecastingmodelsstrong}.

\textbf{Shared Structural Limitation}
Despite their differences, all of the above share a fundamental constraint: $F_\theta$ is optimized independently of the target classification task, and a new $G_\phi$ must be fitted per dataset.
This decoupling has also been identified as a key source of sub-optimality in \citet{shi_trade-off_2022}.\footnote{InceptionTime is an exception to this limitation, as $F_\theta$ and $G_\phi$ are trained jointly on the target dataset.}
Furthermore, $F_\theta$ encodes each series in isolation, without any awareness of the other labeled samples in the dataset, and therefore cannot exploit discriminative structure across classes during encoding.
Most critically, at inference time, $F_\theta$ does not exploit the labeled training samples, and therefore cannot produce representations tailored to the class structure of the dataset at hand.
\timee address these three constraints simultaneously via end-to-end in-context learning.

\subsection{In-Context Learning for Supervised Prediction}
                                                        
In-context learning (ICL) offers an alternative to task-specific classifiers: rather than updating model weights, the model observes $\mathcal{D}_{train}$ at inference and approximates the conditional predictive distribution $p(\cdot \mid x_{test}, \mathcal{D}_{train})$ in a single forward pass.

This idea was first introduced in Prior-data Fitted Networks (PFN)~\citep{muller2024transformersbayesianinference}. TabPFN~\citep{hollmann-iclr23a} extended this paradigm for tabular classification by defining a synthetic Bayesian prior over tasks, while TabPFNv2~\citep{hollmann-nature25a} scaled this substantially by introducing advancement in the synthetic priors and architecture.
Parallel works that adopt the same paradigm include TabICL~\citep{qu2025tabicltabularfoundationmodel} and TabICLv2~\citep{qu2026tabiclv2betterfasterscalable} which introduce different architectures and synthetic priors.

For TSC, \citet{fang_rethinking_2026} introduce ICL by pre-training an ICL head atop a fixed feature encoder. While this yields strong results, it remains structurally two-stage: the encoder is frozen and cannot be shaped by the classification objective. TiCT~\citep{yeh2025tictsyntheticallypretrainedfoundation} addresses the problem more directly by pre-training an end-to-end model on binary tasks, using threshold-based mixing of KernelSynth examples as synthetic prior. \timee extends this line of work to multi-class tasks and grounds the synthetic prior in a structured data-generating process that reflects meaningful class differences in the underlying dynamics. For a more in-depth discussion on the differences in the two approaches, we refer to Appendix~\ref{app:tict-timee-comp}.

\subsection{Synthetic Data Generation for Time Series Foundation Models} \label{sec:synth-data-prior-work}
Due to the scarcity of high-quality labeled real-world time series data, synthetic generate has become a key enabler of zero-shot generalization in many time series foundation models. KernelSynth, originally introduced in Chronos~\citep{ansari_chronos_2024}, generates diverse synthetic time series for forecasting through compositions of Gaussian process kernels. Chronos-2 further introduces using temporal causal models in order to generate synthetic univariate time series.

In a separate direction, several works have proposed data mixing strategies to further diversify synthetic pre-training data. TiRex~\citep{auer2025tirexzeroshotforecastinglong} applies augmentation techniques including spike injection and convex mixing of base sequences, as well as an extended version of KernelSynth. CauKer \citep{xie_cauker_2025} uses Structural Causal Models to produce clustering structure suitable for contrastive pre-training of time series encoders.

ICL pre-training for classification imposes a strictly harder requirement: synthetic data must consist of labeled classification tasks, not merely diverse time series. %
No prior work generates classification tasks in which class identity reflects differences in the data-generating process itself.

\section{Method}\label{sec:method}
We formulate the training objective of \timee and detail the key components of our approach, beginning with the synthetic data generation procedure and concluding with the model architecture and pre-training setup.
\subsection{Problem Formulation and Training Objective}

While existing approaches solve each instance of $p(\cdot \mid \mathbf{x}_{test}, \mathcal{D}_{train})$
independently via fine-tuning or two-stage pipelines, \timee learns an amortized predictor $f_\theta$ that approximates it in a single forward pass with no per-dataset weight updates.
Rather than assuming a fixed dataset or label space, $f_\theta$ accepts tasks of varying support size $n_{train}$ and number of classes $K$.
Following the PFN~\citep{muller2024transformersbayesianinference} training objective, we train $f_\theta$ by meta-learning over synthetic tasks drawn from a prior $p_{synth}$. At each iteration, a synthetic dataset $\mathcal{D}_k = \{(x_i, y_i)\}_{i=1}^N$ is sampled and partitioned into:
\[\mathcal{D}_{k,train} = \{(x_j, y_j)\}_{j=1}^{n_{train}}, \quad \mathcal{D}_{k,test} = \{(x_l, y_l)\}_{l=1}^{n_{test}},
\]
and the model minimizes the negative log-likelihood over query positions:                                                                        
\begin{equation}
    \mathcal{L}(\theta) = -\mathbb{E}_{\mathcal{D} \sim p_{synth}} \left[ \sum_{(x, y) \in \mathcal{D}_{k,test}} \log p_\theta(y \mid x, {D}_{k,train}) \right].
    \label{eq:loss}
\end{equation}

\subsection{Synthetic Data Generation}\label{sec:synth-data}
\begin{figure}
    \centering
    \includegraphics[width=0.9\linewidth]{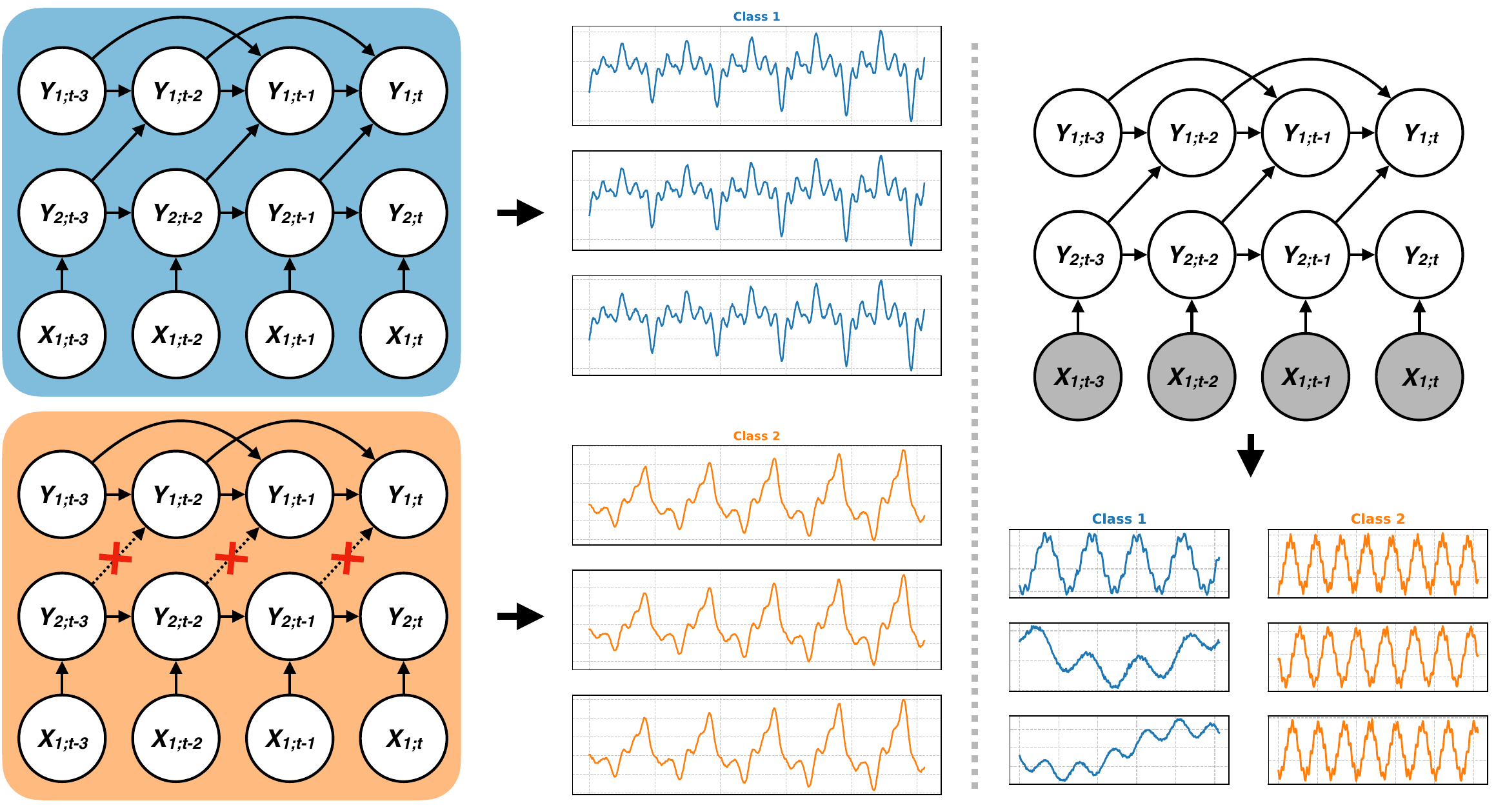}
    \caption{Class generation mechanisms in the \timee prior: \textbf{\textit{Left -- Structural Variation}}: samples from Class 1 (\textcolor{blue-synth}{blue}) are generated under the full graph, while samples from Class 2 (\textcolor{orange-synth}{orange}) are generated under a graph with a subset of edges dropped (\textcolor{red}{$\bm{\times}$}); and \textbf{\textit{Right -- Signal Variation}}: Class 1 (\textcolor{blue-synth}{blue}) and Class 2 (\textcolor{orange-synth}{orange}) share the same underlying signal pattern, and class identity is determined by the warping speed of the exogenous signal (\textcolor{grey-synth}{grey} nodes).}
    \label{fig:tcms}
    \vspace{-1em}
\end{figure}
The design of the prior $p_{synth}$ is central to the performance of \timee. While prior work (Section~\ref{sec:synth-data-prior-work}) has established synthetic data generation as a promising data source for  pre-training time series foundation models, these approaches target either forecasting or produce classification tasks in which class identity does not reflect meaningful differences in the underlying generative process. We address this gap by pre-training \timee on synthetic labeled TSC datasets in which class structure arises from controlled variation in the data-generating process itself.

\textbf{Data-Generating Process.} The foundation of our data-generating process is a temporal structural model over $d$ endogenous variables. Let $\mathbf{y}_t \in \mathbb{R}^d$ denote a multivariate\footnote{The univariate case $d=1$ is also covered by this model.} random variable at time $t$. We model its evolution as a Vector Autoregressive process with exogenous inputs (VARX) \citep{lutkepohl_new_2005}:
\begin{align*}
    \mathbf{y}_t = \sum_{k=1}^{p}\mathbf{A}_k\mathbf{y}_{t-k} + \sum_{j=0}^{s}\mathbf{B}_j{\mathbf{x}_{t-j}} + \bm{\epsilon}_t, \qquad \bm{\epsilon}_t \sim \mathcal{N}(\mathbf{0}, \mathbf{\Sigma})
\end{align*}
where $\mathbf{A}_k \in \mathbb{R}^{d \times d}$ are the lag-$k$ autoregressive coefficient matrices, encoding inter-variable temporal relationships, $\mathbf{x}_t \in \mathbb{R}^q$ is a $q$-dimensional exogenous input signal and $\mathbf{B}_j \in \mathbb{R}^{d \times q}$ maps lag-$j$ exogenous inputs to endogenous variables. 

\paragraph{Class Identity Mechanism.}
Classification structure is introduced through two complementary and independently controllable mechanisms:

\begin{itemize}
    \item \textbf{\textit{Structural Variation:}} Classes are defined by structural variation in the dependency structure encoded in $\mathbf{A}_k$ and $\mathbf{B}_j$ as shown in Figure~\ref{fig:tcms} (left). Concretely, class identity is determined by which the set of edges that are present (or absent) in the dependency graph. This reflects well-established observations that many real-world classification tasks correspond to qualitatively different regimes of inter-variable interaction \citep{qian2025structural}.
    \item \textbf{\textit{Signal Variation.}} Classes are defined by the statistical properties of the exogenous input $\mathbf{x}_t$ while the underlying dependency structure $\mathbf{A}_k$, $\mathbf{B}_k$ remains fixed. In particular, we consider warping of $\mathbf{x}_t$ as class-discriminative feature, as shown in Figure \ref{fig:tcms} (right). The same underlying signal is hereby expressed at different speeds across classes, inducing systematic distributional differences in the endogenous $\mathbf{y}_t$ without any change to the inter-variable structure \citep{kamycki2019data}.
\end{itemize}

\textbf{Augmentations.}
To further enhance the diversity of our synthetic pre-training corpora, we apply a set of augmentations at the individual time series level.
Concretely, we consider three transformation types:
\begin{enumerate*}[(1)]
    \item \textit{smoothing} via a randomized Gaussian filters;
    \item \textit{spike injection}: by sampling spike $s_t$ and augmenting the series by  $\mathbf{y}_t'=\mathbf{y}_t + s_t$ \citep{auer2025tirexzeroshotforecastinglong}; and
    \item \textit{nonlinear transformations}: by applying non-linearity on the time series, i.e. $\mathbf{y}_t' = g(\mathbf{y}_t)$ where $g$ is a randomly sampled nonlinearity.
\end{enumerate*}
Additionally, we apply a mixup scheme~\citep{ansari_chronos_2024,auer2025tirexzeroshotforecastinglong}, to generate convex combinations of synthetic time series across the dataset level.
Full details about the augmentations are provided in Appendix~\ref{app:augmentations} and~\ref{app:mixup}.

\subsection{Model Architecture} \label{sec:model-architecture} %

The architecture of \timee is designed to jointly reason over all time series in the context, enabling in-context comparison between support and query examples in a single forward pass.
It consists of five stages:
(i) \textbf{Tokenization} to represent each series as a sequence of patch embeddings;
(ii) \textbf{Label Conditioning} to inject class information to the patch embeddings;
(ii) \textbf{Encoder Stack} to build class-aware contextual representations for each series
(iv) \textbf{In-Context Reasoning} to perform cross-sample comparison;
(v) \textbf{Decoder} to map query representations into class probabilities.
The complete pipeline is illustrated in Figure~\ref{fig:arch}, with an in-depth evaluation of each component provided in Appendix~\ref{app:model-components}.

\textbf{Tokenization.}
Like many time series foundation models (mentioned in Section~\ref{sec:background}), we first normalize each series to remove the individual scale, in order to encourage the model to focus on temporal shape rather than magnitude.
Like Chronos-2~\citep{ansari_chronos-2_2025}, we apply $sinh^{-1}$ transformation to better handle the outliers.
A subsequent cross-normalization step, computed using only the training-series statistics, further standardizes the representation across the context.
Each series is then divided into non-overlapping patches of size $p = 16$ and projected to $d = 128$ dimensions via a shared linear layer.

\begin{wrapfigure}{r}{\halffig}
   \vspace{-1.5em}
    \includegraphics[width=\linewidth]{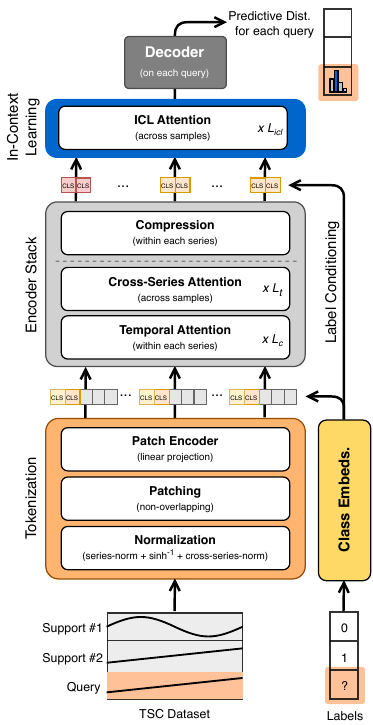}
    \caption{Architecture of \timee.}
    \vspace{-1em}
    \label{fig:arch}
\end{wrapfigure}

\textbf{Label Conditioning.}
A key advantage of the end-to-end paradigm is that class labels are available throughout the forward pass.
We exploit this by conditioning the encoder directly on label identity.
Concretely, following TabICLv2~\citep{qu2026tabiclv2betterfasterscalable}, a learned class embedding is added to each training series' patch representations, while query series receive no such signal.
This allows subsequent attention layers to build class-aware contextual representations that are more discriminative and separable across classes, while preserving the label ambiguity of query patches that the model must resolve through in-context reasoning. 

\textbf{Encoder Stack.}
The goal of the encoder is to produce a compact, class-aware summary of each series that captures both its internal temporal structure and its relationship to other series in the context.
To this end, $K$ learnable \texttt{CLS} tokens are prepended to each series' patch sequence, yielding representations $\mathbf{Z} \in \mathbb{R}^{N \times (P+K) \times d}$.
The encoder then operates in two sequential phases: $L_t$ \emph{temporal} attention layers first process each series independently across its own patches using Rotary Positional Embeddings (RoPE)~\citep{su2021roformer}, capturing within-series temporal structure; $L_c$ \emph{cross-series} attention layers then apply attention across all $N$ series at each patch position, allowing each series to condition on the full labeled context.
Queries attend to support examples but not to one another (see Appendix~\ref{app:masking} for details).
Finally, a compression layer collapses each series into its $K$ \texttt{CLS} vectors by attending only over patch positions, producing a compact series summary $\mathbf{r}_i \in \mathbb{R}^{K \times d}$.

\textbf{In-Context Reasoning.}
Given compressed summaries of all series, the in-context reasoning layers perform the classification decision by comparing each query against the labeled support examples in the context.
The $N$ series summaries $\{\mathbf{r}_i\}$ are processed by $L_\mathrm{icl} = 2$ transformer layers operating in a 512-dimensional space.
Class embeddings are again added to training summaries to keep support representations class-aware at this stage.
The same masking strategy applies (Appendix~\ref{app:masking}), ensuring queries attend over labeled support examples without conditioning on one another.

\textbf{Decoder.}
To produce a final classification decision, query representations are passed through layer norm and a two-layer MLP to produce logits over $C$ classes.
The model is trained with cross-entropy loss over query positions (Equation~\ref{eq:loss}).

\textbf{Many-Class Extension.} \label{sec:many-class-ovr}
\timee is trained with at most $C_\mathrm{max} = 10$ classes, which covers the majority of UCR datasets.
For datasets with $C > C_\mathrm{max}$, we adopt a one-vs-rest (OvR) decomposition strategy, reducing any $C$-class problem to $C$ binary classification contexts.
While more sophisticated strategies exist, we find OvR to be a simple and effective solution; we leave richer many-class handling as future work.

\subsection{Pretraining and Inference}\label{sec:pre-training-inference}
\textbf{Pretraining Setup.}
\timee is pretrained on 7M synthetic TSC tasks, derived from 1.5M unique VARX-generated datasets via augmentation (Section~\ref{sec:synth-data}).
Each task is constructed by uniformly sampling a support-query split and cropping all series in the context to a shared length sampled uniformly from $[16, 1024]$.
We train for $30K$ steps on a single NVIDIA H200 GPU with \texttt{bf16} automatic mixed precision, taking approximately $40$ hours.
Following recent advances in optimizer design, we use Muon~\citep{jordan2024muon} for matrix parameters and AdamW~\citep{loshchilov2019decoupledweightdecayregularization} for all remaining weights, which we find leads to faster and more stable convergence than AdamW alone.
Details of pre-training setup are included in Appendix~\ref{app:pretraining}.

\textbf{Inference.}
To optimize inference, \timee ensembles predictions across a small set of differently preprocessed inputs, combining two elementary transformations. We apply interpolation of short time series and first-order differencing, both individually and in composition. Final predictions are obtained by averaging predicted distributions across 4 ensemble members. We provide a more detailed description of our inference pipeline in Appendix~\ref{app:inference-pipeline}.

\section{Experiments} \label{sec:experiments}

\begin{wrapfigure}{r}{\halffig}
    \vspace{-3.5em}
    \includegraphics[width=\halffig]{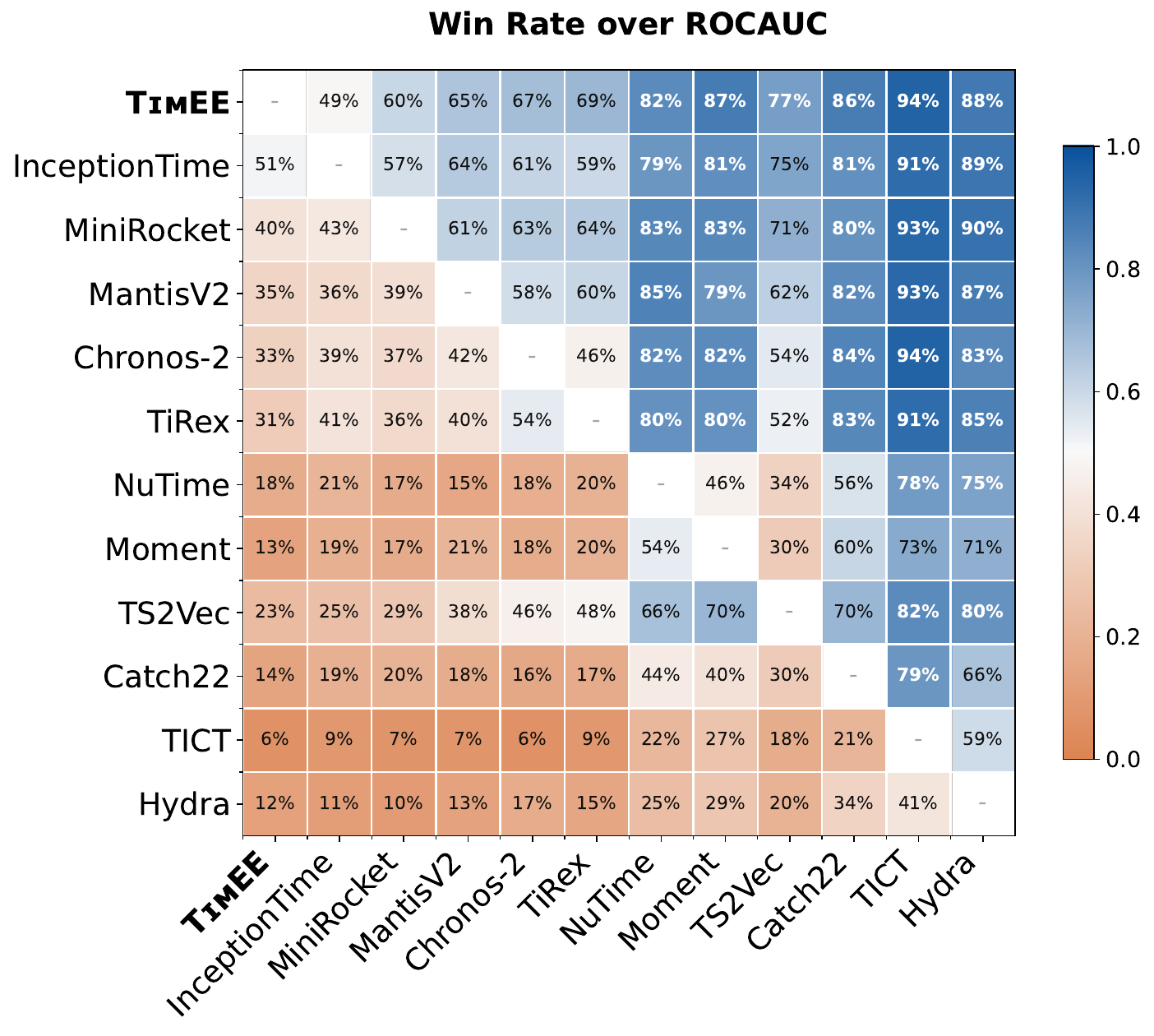}
    \vspace{-1.5em}
    \caption{Win rate over \textsc{ROC AUC} across 128 UCR benchmark datasets. \timee ranks joint first alongside \textsc{InceptionTime}, outperforming all other compared methods. Notably, \timee outperforms all existing pre-trained foundation models by a significant margin.}
    \vspace{-4.5em}
    \label{fig:winrate}
\end{wrapfigure}

We evaluate \timee against a broad set of state-of-the-art baselines for time series classification, assessing predictive performance, calibration, and inference efficiency.

\subsection{Evaluation setup}
\textbf{Datasets.}
We evaluate \timee on the UCR archive~\citep{chen_ucr_2015}, a standard benchmark comprising 128 univariate TSC datasets, using the data splits and loading pipeline provided by
\textit{aeon}~\citep{aeon}\footnotemark. 

\textbf{Baselines.}
We compare against a broad set of baselines spanning three categories (Section~\ref{sec:background}): classical and feature-based methods (DTW-1NN, Catch22, MiniRocket, Hydra), deep learning methods (InceptionTime, TSVec), and time series foundation models (NuTime, MOMENT, MantisV2, TiCT, Chronos-2, and TiRex).

As discussed in Section~\ref{sec:two-stage-paradigm}, most foundation model baselines require tuning---such as selecting the optimal intermediate layer representation---to achieve competitive performance;
we therefore report both default and tuned results for these models.
Full tuning details are provided in Appendix~\ref{app:baselines}.

Regarding downstream classifier, for Catch22, MiniRocket, and Hydra, we use the \textit{aeon} implementations 
with their canonical classifier pairings. For the remaining representation and foundation models, we select between Random Forest and SVM based on validation performance, with Random Forest generally preferred except for TS2Vec where SVM performs better. See Appendix~\ref{app:baselines} for more details.

\subsection{Main Results}

\begin{figure}
    \centering
    \begin{minipage}{0.42\textwidth}
        \centering
        \includegraphics[width=0.97\linewidth]{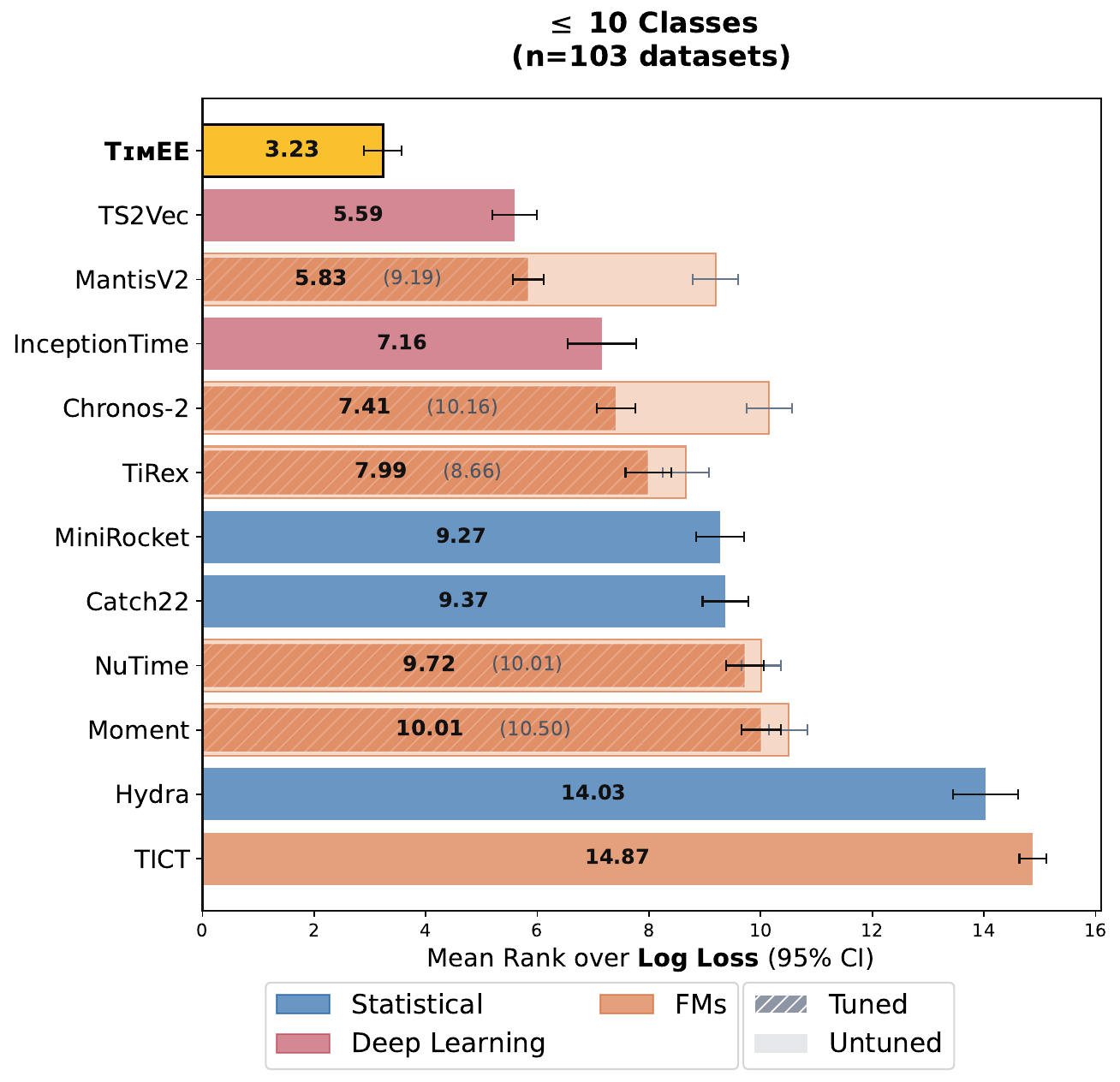}
        \caption{Mean rank on log loss across UCR datasets with maximum 10 classes ($n=103$). \timee achieves the lowest mean rank, indicating well-calibrated predictive distributions across datasets.}
        \label{fig:log-loss-10}
    \end{minipage}
    \hfill %
    \begin{minipage}{0.55\textwidth}
        \centering
        \includegraphics[width=1.0\linewidth]{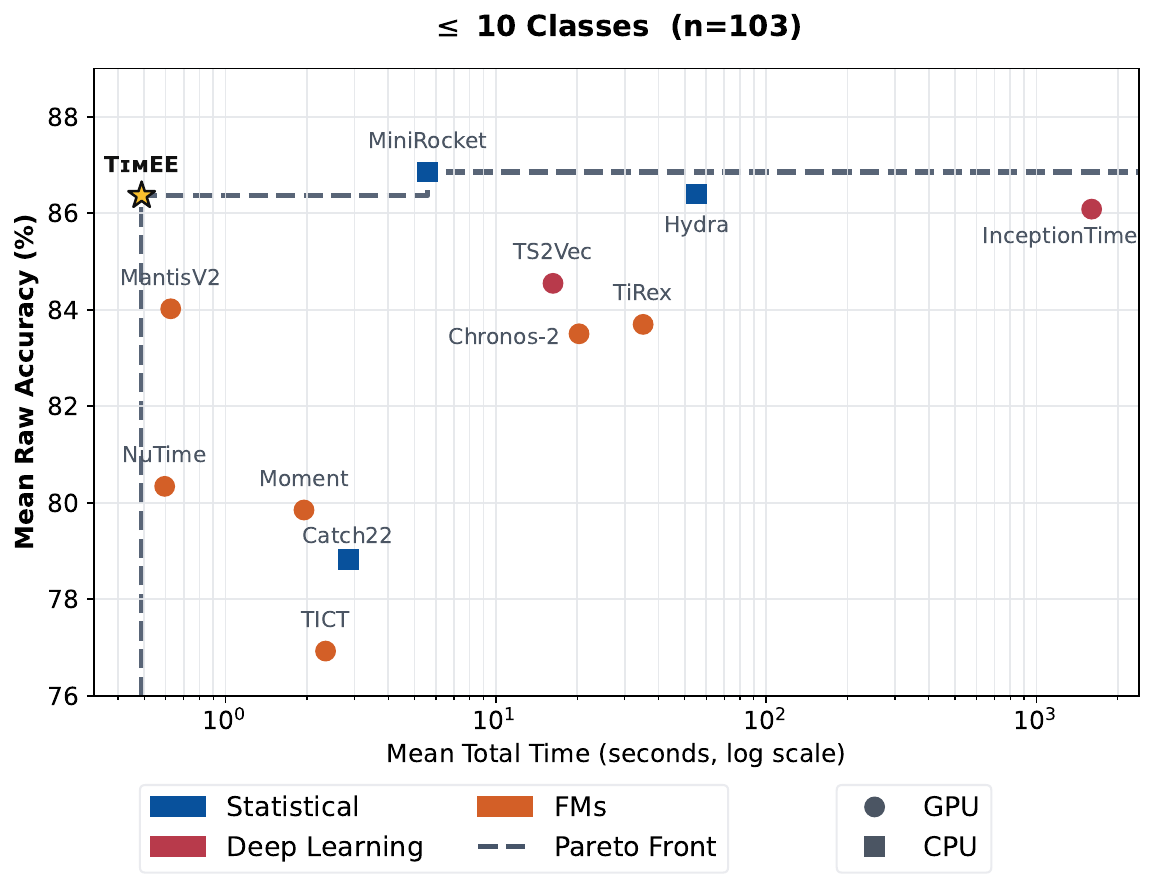}
        \vspace{-1.5em}

        \caption{Pareto front of inference speed vs. ROC AUC on UCR datasets with at most $10$ classes. \timee lies on the Pareto front, approaching the performance of \textsc{MiniRocket} with 15-20x faster inference.}
        \label{fig:pareto}
    \end{minipage}
\end{figure}

Figure~\ref{fig:main-res} illustrates the mean rank across 128 UCR datasets. \timee achieves the best mean rank on ROC AUC, outperforming all baselines in class discrimination. While \timee ranks 3rd in Accuracy, its superior ROC AUC suggests it is the most robust choice across varying deployment thresholds and class distributions \citep{10.5555/645527.657469}. Notably, ridge-regression-based models like \textsc{Hydra} and \textsc{MiniRocket} rank substantially lower on ROC AUC despite competitive accuracy, a discrepancy indicative of overconfident predictions that degrade outside of a fixed 0.5 decision threshold. We present full per-dataset results in Appendix~\ref{app:per-dataset-res}%

\textbf{Probabilistic Calibration.}
To evaluate whether these scores represent well-calibrated probabilities, we report mean rank over log-loss for datasets with $\leq10$ classes\footnote{We restrict this to $\leq10$ classes as larger sets require OvR decomposition for \timee, which introduces normalization artifacts that would confound comparison against native multi-class baselines. We provide a comparison for $>10$ classes in Appendix~\ref{app:inference-time}} ($n=102$) in Figure~\ref{fig:log-loss-10}. \timee achieves the top rank by a significant margin. As Log Loss is a strictly proper scoring rule, this result confirms that \timee provides better-calibrated probabilistic estimates. The gap in Accuracy suggests that while ridge-based methods may force correct hard-label decisions on boundary samples via their objective function, \timee maintains a more reliable representation of uncertainty, yielding more robust inputs for downstream decision-making.

\begin{wrapfigure}{r}{\halffig}
    \vspace{-2em}
    \includegraphics[width=\halffig]{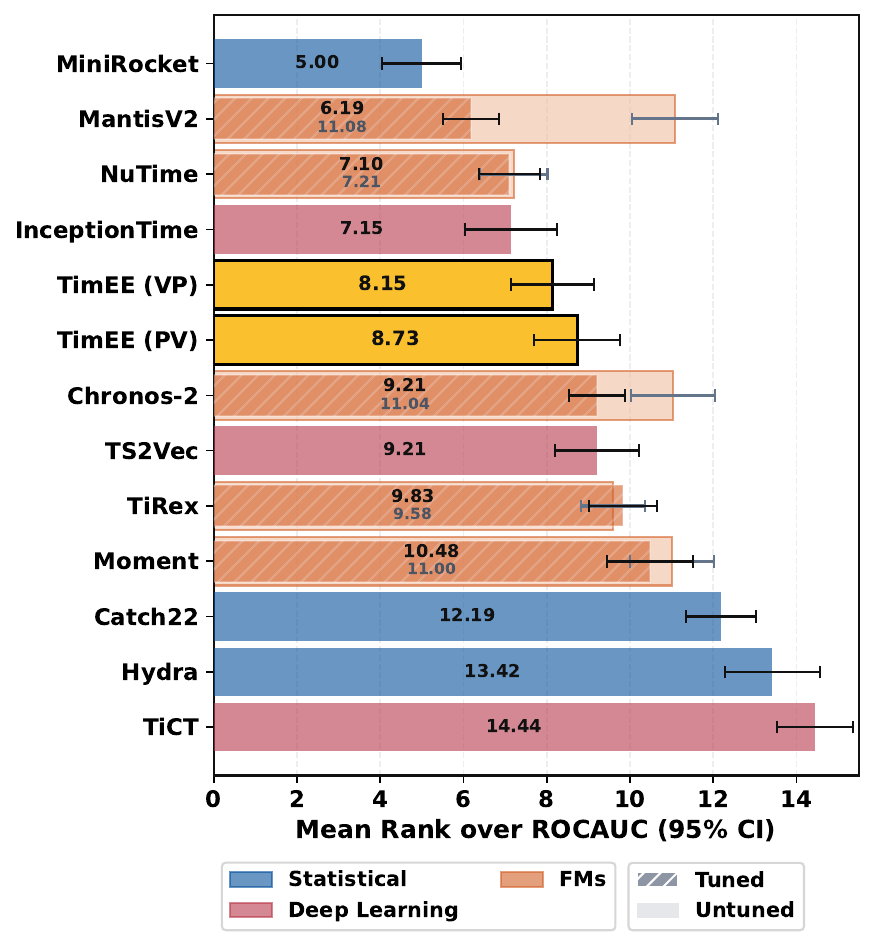}
    \caption{Mean rank over ROC AUC on 24 UEA datasets.
    We compare two variants of \timee multivariate extensions: (i) \timee (PV) denotes \textit{per-variate} variant; (ii) \timee (VP) denotes \textit{variate-pooling} variant.
    \timee reaches competitive results despite being primary developed for univariate setting.
    }
    \label{fig:multivariate}
    \vspace{-5em}
\end{wrapfigure}

\textbf{Inference Speed.}
\timee's end-to-end ICL mechanism enables fast inference without any per-dataset training. While many other methods require a costly separate training phase, \timee combines the traditional fit+predict stages in a single forward pass that directly produces predictions. 
As a result of this, as shown in Figure~\ref{fig:pareto}, \timee lies on the Pareto front of inference speed vs. predictive performance.
Note that this comparison is restricted to datasets with at most 10 classes, as OvR is required for larger class counts and this introduces additional inference overhead (see Appendix Figure~\ref{fig:pareto-front-more-than-10}).
This remains as a limitation of \timee and is discussed in Section~\ref{sec:limitation}.

\subsection{Ablations} \label{sec:ablations}

We present ablation studies on: \begin{enumerate*}
    \item examining how \timee can be extended to multivariate tasks; and
    \item probing the internal mechanisms that drive its in-context classification performance
\end{enumerate*}.

\paragraph{Multivariate Extension.}\label{sec:abl-mv}
We consider two strategies for adapting \timee to multivariate settings:
\begin{enumerate*}[label=(\roman*)] \item \textit{Per-variate}: predict independently on each variate and average the resulting class probabilities; and
    \item \textit{Variate-pooling} (details in Appendix~\ref{app:extension-to-mv}): encode each variate independently and aggregate the $M$ resulting \texttt{CLS} token sets via attention pooling before the ICL phase.
\end{enumerate*}
The \textit{variate pooling} variant is fine-tuned on synthetic multivariate data generated with our prior (Section~\ref{sec:synth-data}).
We evaluate both strategies against the baselines from Section~\ref{sec:experiments} on the UEA archive.
We restrict to 24 out of 30 UEA datasets\footnote{SpokenArabicDigits, InsectWingbeat, PhonemeSpectra, FaceDetection, EigenWorms, and PenDigits are excluded.} as baselines' prediction failure.
As shown in Figure~\ref{fig:multivariate}, the \textit{per-variate} variant achieve competitive performance despite treating variates independently.
Fine-tuning the \textit{variate-pooling} variant, however, only yields a marginal gain.
We attribute this to a limitation of our synthetic data generation: although the prior is natively extensible to multivariate inputs, the per-variate discriminative signal appears sufficient in practice, leaving little room for cross-variate aggregation to help. We discuss this further in Appendix~\ref{app:mv-synth}.

\begin{figure}
    \centering
    \includegraphics[width=1.0\linewidth]{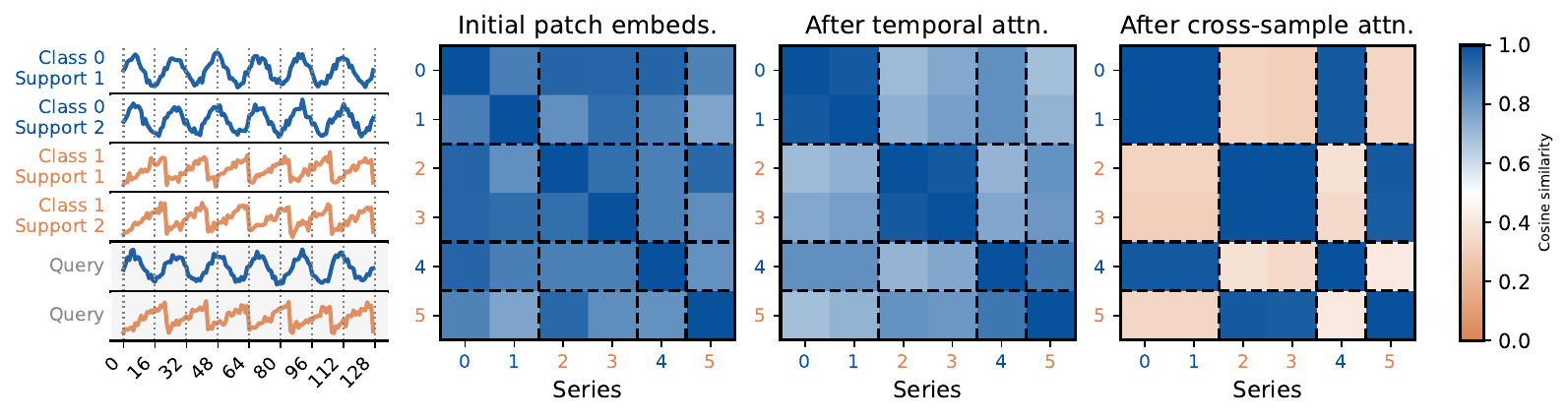}
    \caption{
        Representation alignment across encoder stages.
        The leftmost panel shows six time series from a 2-shot binary classification task on synthetic data: class \textit{0} (blue, sinusoidal) and class \textit{1} (orange, sawtooth).
        The three heatmaps show pairwise cosine similarity of series representations at successive stages of the encoder, comparing the patch embeddings at (i) initial state; (ii) after temporal attention; (iii) after cross-series attention.} %
    \label{fig:attention}
    \vspace{-1em}
\end{figure}
\paragraph{How the Encoder Stack Builds Class-Discriminative Representations.}
To understand how the encoder progressively builds class-discriminative representations, Figure~\ref{fig:attention} traces pairwise cosine similarities across its successive stages. Initially, patch embeddings are largely undifferentiated -- similarities are uniformly high across all series regardless of class membership.
Temporal attention begins to differentiate series by refining within-series structure, and same-class representations grow slightly more similar to one another.
Cross-series attention then sharply accentuates this emerging structure akin to a contrastive separation in representation space.
Crucially, query series --- which receive no class-label signal during the forward pass --- nonetheless align to their correct class cluster, confirming that the encoder builds class-discriminative representations that prime the subsequent in-context reasoning stage.

\section{Conclusion} \label{sec:limitation}

We introduced \timee, a foundation model for time series classification 
that departs from the two-stage paradigm by framing TSC as an in-context learning 
problem.
The key enabler is a synthetic prior grounded in a structured  data-generating process, which produces labeled classification tasks where class identity reflects meaningful differences in the underlying dynamics.
To our knowledge, no prior work addresses this for TSC.
Our strong empirical results on the 128 UCR datasets suggest that the gap between the two-stage paradigm and end-to-end 
ICL is not merely from architecture, but also from the right synthetic prior and the ability to 
leverage class structure at inference time.
We hope this work encourages further exploration of prior design, model scale, and richer class generation mechanisms, which is likely to lead to further substantial performance improvements.

\textbf{Limitation 1: Many-class handling.} \timee is pre-trained with at most 10 
classes and relies on a one-vs-rest decomposition for larger label spaces, which 
introduces significant inference overhead and diminishes the speed advantage of 
ICL-based classification (Figure~\ref{fig:pareto-large}). Extending \timee 
to natively support larger class counts is an important direction for future work.

\textbf{Limitation 2: Multivariate classification.} Extending \timee to multivariate 
inputs via variate pooling yields only marginal gains over the simpler per-variate 
baseline, which we attribute to a limitation of the VARX prior: class-discriminative 
information tends to concentrate in individual variates, leaving little signal for 
cross-variate aggregation to exploit. Designing priors that better leverage 
inter-variate structure remains an open problem.

\textbf{Future Direction: Prior design and scale.} \timee is a small model trained on a 
single family of synthetic priors. The strong UCR results suggest that prior quality 
is a primary lever for performance in this paradigm --- richer class generation 
mechanisms and larger model scale are therefore natural and promising avenues for 
future work.

\newpage

\section*{Acknowledgments and Disclosure of Funding}
J.K. is supported by the Konrad Zuse School of Excellence in Learning and Intelligent Systems (ELIZA) through the DAAD program Konrad Zuse Schools of Excellence in Artificial Intelligence, sponsored by the Federal Ministry of Education and Research. L.P. acknowledges funding by the Deutsche Forschungsgemeinschaft (DFG, German Research Foundation) under SFB 1597 (SmallData), grant number 499552394;
F.H. acknowledges the financial support of the Hector Foundation.
The authors gratefully acknowledge the computing time made available to them on the high-performance computer at the NHR Center of TU Dresden. This center is jointly supported by the Federal Ministry of Research, Technology and Space of Germany and the state governments participating in the NHR (\url{www.nhr-verein.de/unsere-partner}). We thank the maintainers of the UCR \citep{chen_ucr_2015} and UEA \citep{bagnall2018ueamultivariatetimeseries} archives for providing and maintaining the datasets used for evaluation. We thank the developers of open-source libraries used in the development of \timee, including but not limited
to: \texttt{torch} \citep{pytorch}, \texttt{transformers} \citep{wolf-etal-2020-transformers}, \texttt{scikit-learn} \citep{scikit-learn}, \texttt{numpy} \citep{numpy} and \texttt{aeon} \citep{aeon}.

\section*{Author Contributions}
J.K. and S.B.H. co-led the project, code development and experiments and writing of the paper. J.K. led the synthetic data generation and contributed to the model development, training infrastructure, and evaluation.
S.B.H. led the model development and contributed to the synthetic data generation, training infrastructure, and evaluation.
M.M. contributed to the evaluation and writing of the paper.
F.H. supervised L.P. during the project.
L.P. supervised J.K., S.B.H., and M.M. during the project, contributed ideas, and technical advice.

\section*{Competing Interests}
L.P., and F.H. are affiliated with Prior Labs, a company
focused on developing tabular foundation models. The authors declare no further competing interests.

\bibliography{lib,proc,strings,TimeSeriesClassification}
\bibliographystyle{plainnat}

\clearpage
\appendix

\section{Related Work Details}

\subsection{Comparison to related methods for Time Series Classification}

\definecolor{cmarkcolor}{RGB}{40,120,50}
\definecolor{xmarkcolor}{RGB}{180,40,40}
\newcommand{\cmark}{\textcolor{cmarkcolor}{\checkmark}}
\newcommand{\xmark}{\textcolor{xmarkcolor}{$\times$}}

\begin{table}[ht]
\caption{%
  Structural decomposition of compared methods into encoder $F_\theta$ and
  classifier $G_\phi$.
  \emph{New $G_\phi$} indicates whether a task-specific classifier must be
  re-fitted for each new dataset.
  \emph{Class-aware} indicates whether $F_\theta$ has access to the labeled
  support set at inference time.
  $^\dagger$Competitive performance requires tuning the intermediate layer used
  for feature extraction~\citep{auer2025pretrainedforecastingmodelsstrong}).
}
\label{tab:two-stage}
\centering
\resizebox{\linewidth}{!}{%
\setlength{\tabcolsep}{6pt}
\renewcommand{\arraystretch}{1.3}
    \begin{tabular}{
  l
  p{4cm}
  p{2cm}
  p{2cm}
  p{2cm}
  c
  c
}
\toprule
\textbf{Method}
  & \textbf{Encoder $F_\theta$}
  & \textbf{$F_\theta$ objective}
  & \textbf{$F_\theta$ data}
  & \textbf{Classifier $G_\phi$}
  & \textbf{\begin{tabular}{@{}c@{}}New $G_\phi$\\per dataset\end{tabular}}
  & \textbf{\begin{tabular}{@{}c@{}}Class-aware\\at inference\end{tabular}}
\\
\midrule
\multicolumn{7}{l}{\textit{Classical \& Time Series Features Engineering Methods}} \\
\midrule
DTW-1NN
  & Dynamic warping distance
  & ---
  & ---
  & 1-NN
  & ---
  & \xmark
\\
Catch22
  & 22 hand-crafted statistics
  & ---
  & ---
  & Linear / SVM
  & \cmark
  & \xmark
\\
Rocket / MiniRocket
  & Random conv.\ kernels; PPV \& max pooling
  & ---
  & ---
  & Ridge regression
  & \cmark
  & \xmark
\\
Hydra
  & Competing random conv.\ kernels; per-group argmax counts
  & ---
  & ---
  & Ridge regression
  & \cmark
  & \xmark
\\[4pt]

\midrule
\multicolumn{7}{l}{\textit{Per-dataset deep learning}} \\
\midrule
InceptionTime
  & Multi-scale conv.\ encoder
  & Cross-entropy (joint)
  & Target dataset
  & Softmax head (joint)
  & \cmark
  & \xmark
\\
TS2Vec
  & Hierarchical contrastive encoder
  & Contrastive (unsupervised)
  & Target dataset (unlabeled)
  & Linear / SVM
  & \cmark
  & \xmark
\\[4pt]

\midrule
\multicolumn{7}{l}{\textit{Foundation models --- TSC-focused}} \\
\midrule
NuTime
  & Transformer; multi-scaled window embedding
  & BYOL contrastive
  & ${>}1$M seq.
  & Linear / SVM
  & \cmark
  & \xmark
\\
MOMENT
  & Masked autoencoder transformer
  & Masked reconstruction
  & Large real-TS corpus
  & Linear / SVM$^\dagger$
  & \cmark
  & \xmark
\\
Mantis
  & ViT encoder
  & Contrastive
  & Large real-TS corpus
  & Linear / SVM$^\dagger$
  & \cmark
  & \xmark
\\
MantisV2
  & ViT encoder + test-time strategies
  & Contrastive
  & CauKer synthetic
  & Linear / SVM$^\dagger$
  & \cmark
  & \xmark
\\[4pt]

\midrule
\multicolumn{7}{l}{\textit{Foundation models --- forecasting, repurposed}} \\
\midrule
Chronos-2
  & Transformer encoder
  & Forecasting (quantile)
  & Large TS corpus
  & Linear / SVM$^\dagger$
  & \cmark
  & \xmark
\\
TiRex
  & Transformer encoder
  & Forecasting
  & Large TS corpus
  & Linear / SVM$^\dagger$
  & \cmark
  & \xmark
\\[4pt]

\midrule
\multicolumn{7}{l}{\textit{End-to-end in-context learning}} \\
\midrule
TiCT
  & ResNet encoder
  & ICL cross-entropy (binary)
  & KernelSynth mixup
  & ICL head (bit-encoded labels)
  & \xmark
  & \cmark
\\
\textbf{TimEE} (ours)
  & Transformer encoder
  & ICL cross-entropy
  & VARX structured synthetic
  & ICL head (single forward pass)
  & \xmark
  & \cmark
\\

\bottomrule
\end{tabular}

}%
\end{table}

\subsection{Comparison \timee vs TiCT~\citep{yeh2025tictsyntheticallypretrainedfoundation}}\label{app:tict-timee-comp}

Concurrent with this work, TiCT proposes end-to-end ICL for TSC with synthetic pre-training --- the paradigm most directly related to ours.
TiCT pre-trains a ResNet-based encoder on binary synthetic tasks via KernelSynth mixup, uses bit-based label encoding for arbitrary class counts, and selects context via 64-nearest-neighbor retrieval at inference. TimEE departs from TiCT in three key respects: our class generation mechanism --- structural breaks in a VARX process -- produces multi-class tasks where class identity reflects differences in the underlying dynamics; our transformer architecture explicitly models both temporal within-series structure and cross-series ICL; and we use the full training set as context without a retrieval step. We evaluate against substantially stronger baselines on both UCR and UEA. %
That TiCT and TimEE independently converge on the same paradigm strengthens the case that end-to-end ICL with synthetic pre-training is a promising direction for TSC.

\section{Model Architecture}
\subsection{Model Hyperparameters}\label{app:model-hps}
We report the model hyperparameters for \timee in Table~\ref{tab:architecture}
\begin{table}[h]
\centering
\caption{Architecture hyperparameters of \timee.}
\label{tab:architecture}
\begin{tabular}{lc}
\toprule
\textbf{Hyperparameter} & \textbf{Value} \\
\midrule
\multicolumn{2}{l}{\textit{Tokenization}} \\
\\
Patch size & $16$ \\
Patch stride & $16$ \\
\midrule
\multicolumn{2}{l}{\textit{Architecture}} \\
\\

Model dimension $d_{\text{model}}$ & $128$ \\
Temporal layers, $L_t$ & $5$ \\
Cross-series layers, $L_c$ & $5$ \\
ICL layers, $L_{icl}$ & $2$ \\
Attention heads & $4$ \\
$d_{kv}$ & $32$ \\
$d_{ICL_{kv}}$ & $64$ \\
MLP hidden dimension & $512$ \\
Decoder MLP hidden dimension & $512$ \\
Dropout & $0.1$ \\

\midrule
\multicolumn{2}{l}{\textit{Classification}} \\
\\
CLS tokens & $4$ \\
Maximum classes & $10$ \\
\bottomrule
\end{tabular}
\end{table}

\subsection{Attention Masking Strategy}
\label{app:masking}

Let $\mathcal{S} = \{1, \ldots, n\}$ and $\mathcal{Q} = \{n+1, \ldots, n+m\}$ denote the indices of training and query series respectively, with $N = n + m$.
The binary attention mask $\mathbf{M} \in \{0, 1\}^{N \times N}$, applied consistently across all cross-series attention layers, is defined as:

\begin{equation}
    \mathbf{M}_{ij} = \begin{cases}
        1 & \text{if } i \in \mathcal{S} \\
        1 & \text{if } i \in \mathcal{Q} \text{ and } j \in \mathcal{S} \cup \{i\} \\
        0 & \text{otherwise}
    \end{cases}
\end{equation}

where $\mathbf{M}_{ij} = 1$ indicates that series $i$ is permitted to attend to series $j$, and masked positions are set to $-\infty$ in the pre-softmax attention logits.
The resulting block structure is:

\begin{equation}
\mathbf{M} = \bordermatrix{
    & \mathcal{S} & \mathcal{Q} \cr
    \mathcal{S} & \cmark & \cmark \cr
    \mathcal{Q} & \cmark & \mathbf{I} \cr
}
\end{equation}

Training series attend to all series; query series attend to all training series and themselves, but not to other query series.
Temporal attention layers and the compression layer operate within each series independently and require no masking.

\subsection{Extension to Multivariate} \label{app:extension-to-mv}

\textbf{Architecture.}
\timee is designed and trained to operate on univariate series only.
To handle multivariate inputs $\mathbf{X} \in \mathbb{R}^{N \times S \times M}$, we encode each of the $M$ variates independently through the shared encoder, producing $M$ sets of \texttt{CLS} token vectors per series. 
These are aggregated via \emph{slot-wise attention pooling}: each of the $K$ \texttt{CLS} slots has a learnable query that attends over the $M$ variate embeddings via a softmax, allowing each slot to specialize on the most informative variates.
The resulting pooled summary $\mathbf{r}_i \in \mathbb{R}^{K \cdot d}$ is then passed to the ICL phase.      
  
\textbf{Fine-tuning Details.}
Queries are zero-initialized so that pooling starts as a uniform average, providing a stable initialization from the pretrained univariate checkpoint.
The encoder weights are frozen and only the pooling queries, ICL block, and decoder are updated during multivariate fine-tuning.

\subsection{Ablation of Model Components.}\label{app:model-components}
We conduct the following ablation studies in the following training setup while we modify the architecture configuration by:
\begin{itemize}
    \item scaling down the model configuration to $d_{kv}=16$, $n_{heads}=2$;
    \item fixing the sequence length at 512 with batch size 64; 
    \item training on our synthetic univariate data for 25K steps;
    \item evaluating on the train split of 86 UCR datasets (defined by: at least 100 samples, the least frequent classes must at least have 2 examples) with stratified 3-fold cross-validation.
\end{itemize}
Results reported are averaged over 3 seeds.

\textbf{Contribution of Temporal and Cross-Series Attention.}
    We ablate the contribution of cross-series (vertical) attention within the encoder by varying the allocation of encoder layers between temporal ($L_t$) and cross-series ($L_c$) attention, keeping the total fixed at $L_t + L_c = 8$ with $L_\mathrm{icl} = 2$.
    
    \begin{table}[h]
    \centering      
    \small
    \caption{Ablation of encoder layer allocation between temporal and cross-series attention.}
    \begin{tabular}{lcccc}
    \toprule
    Configuration & $L_t$ & $L_c$ & \textsc{Roc Auc} & Accuracy (\%) \\
    \midrule
    Temporal only    & 8 & 0 & $70.70 \pm 0.08$ & $78.21 \pm 0.39$ \\
                   & 6 & 2 & $71.06 \pm 0.10$ & $79.96 \pm 0.29$ \\
    Base config  & 4 & 4 & $71.26 \pm 0.09$ & $80.32 \pm 0.57$\\                          
                   & 2 & 6 & $71.20 \pm 0.05$ & $80.45 \pm 0.36$\\
    Cross-series only & 0 & 8 & $71.25\pm0.02$ & $79.65 \pm 0.15$\\
    \bottomrule
    \end{tabular}
    \label{tab:ablation-encoder}
    \end{table}
    
    Removing cross-series attention entirely ($L_c = 0$) causes a clear 2pp drop, confirming that encoder-level cross-series context provides complementary signal to the ICL phase and cannot be recovered from \texttt{CLS}-level comparison alone.
    
    Removing temporal attention ($L_t = 0$) causes a smaller but still notable drop of approximately 1pp; local patch features provide a partial substitute for temporal structure, while cross-series comparison at the patch level is harder to replicate from compressed representations.
    
    Performance is stable across all mixed configurations, indicating that the model is insensitive to the exact allocation as long as both components are present.
    
    We note that all configurations retain a single compression attention layer (Section~\ref{sec:model-architecture}), so the $L_t = 0$ condition is not strictly temporal-attention-free.

\textbf{Contribution of conditioning on Class Label.}                                                                                                             
  We ablate the two class embedding injection sites: patch-level conditioning (added to training patches before the encoder) and ICL-level conditioning (added to \texttt{CLS} tokens before the ICL block).

   \begin{table}[h]
   \centering                                                                                                          
  \small
  \caption{Ablation of label conditioning sites.}
  
  \begin{tabular}{cccc}
  \toprule                                                                                                            
  Patch conditioning & ICL conditioning & \textsc{Roc Auc} & Accuracy (\%) \\
  \midrule
  \checkmark & \checkmark & $71.26 \pm 0.09$ & $80.32 \pm 0.57$ \\
  \checkmark & -- & $71.16 \pm 0.08$ & $80.35 \pm 0.13$\\
  -- & \checkmark & $71.01 \pm 0.09$ & $78.55 \pm 0.18$ \\                                                                   
  -- & -- & $50.13 \pm 0.43$  & $22.27 \pm 0.37$ \\                                                                                    
  \bottomrule     
  \end{tabular}                                                                                                       
  \label{tab:ablation-conditioning}                                                                                   
  \end{table}
  
  Without any class conditioning, accuracy collapses to below 30\%: the model has no access to label information during the forward pass and cannot associate support series with their classes.
Patch-level conditioning is the dominant contributor—and in isolation 
sufficient: removing ICL-level conditioning while retaining patch-level 
conditioning leaves accuracy essentially unchanged 
($80.32 \to 80.35$, within run-to-run variance), 
whereas retaining only ICL-level conditioning drops accuracy by $\sim$1.8pp 
($80.32 \to 78.55$), because the encoder's cross-series attention operates 
on class-unaware patch representations and produces less discriminative 
\texttt{CLS} summaries.

Although ICL-level conditioning provides no measurable benefit over 
patch-level conditioning alone, we retain it in our final model: it was 
included in the large-scale pretraining run, and the negligible overhead 
does not justify re-training without it.
Together, the two sites interact non-linearly: their individual 
contributions do not account for the full ${\sim}$50pp collapse when both 
are removed, confirming that the model requires at least one explicit 
class signal to function.

\newpage

\section{Synthetic Data}\label{app:synth-data}
\subsection{Details on the \timee Prior}
\subsubsection{Sampling Procedure}
Synthetic datasets under the structural variation mechanism are sampled from a VARX-based prior \citep{lutkepohl_new_2005}, in which class identity is determined by class-specific edge dropout applied to a shared base dependency graph. The full sampling procedure is given in Algorithm~\ref{alg:structural-variation-prior}.
Synthetic datasets under the signal variation mechanism are sampled from a VARX-based prior in which the dependency structure $\mathbf{A}_k, \mathbf{B}_j$ remains fixed across all classes, and class identity is instead determined by statistical indicators of the exogenous input signal. The full sampling procedure is given in Algorithm~\ref{alg:signal-variation-prior}.

\begin{algorithm}[H]
\caption{Structural Variation Prior}
\label{alg:structural-variation-prior}
\begin{algorithmic}[1]
\Require Number of classes $C$, Number of samples per class $n$, Max lag orders $p_{\max}$, $s_{\max}$
\State Sample dimensions $d \sim \mathcal{U}\{d_{\min}, d_{\max}\}$, $q \sim \mathcal{U}\{1, q_{\max}\}$
\For{$c = 1, \dots, C$}
    \State Sample lag orders $p \sim \mathcal{U}\{1, p_{\max}\}$ and $s \sim \mathcal{U}\{0, s_{\max}\}$
    \State Sample sparse base coefficient matrices $\mathbf{A}_k^{(c)} \in \mathbb{R}^{d \times d}$ for $k = 1, \dots, p$ 
    \State Sample $\mathbf{B}_j^{(c)} \in \mathbb{R}^{d \times q}$ for $j = 0, \dots, s$
    \For{$i = 1, \dots, n$}
        \State Sample exogenous signal $\mathbf{x}_t \in \mathbb{R}^q$ for $t = 1, \dots, T$
        \State Unroll VARX using $\mathbf{A}_k^{(c)}, \mathbf{B}_j^{(c)}$ to obtain $\mathbf{y}_1, \dots, \mathbf{y}_T$
        \State Assign class label $y_i \leftarrow c$
    \EndFor
\EndFor
\State \Return $\{(\mathbf{y}_i^{(1:T)}, y_i)\}$
\end{algorithmic}
\end{algorithm}

\begin{algorithm}[H]
\caption{Signal Variation Prior}
\label{alg:signal-variation-prior}
\begin{algorithmic}[1]
\Require Number of classes $C$, Number of samples per class $n$, Max lag orders $p_{\max}$, $s_{\max}$, Signal variation range $[\omega_{min}, \omega_{max}]$
\State Sample dimensions $d \sim \mathcal{U}\{d_{\min}, d_{\max}\}$, $q \sim \mathcal{U}\{1, q_{\max}\}$
\For{$c = 1, \dots, C$}
    \State Sample lag orders $p \sim \mathcal{U}\{1, p_{\max}\}$ and $s \sim \mathcal{U}\{0, s_{\max}\}$
    \State Sample sparse base coefficient matrices $\mathbf{A}_k^{(c)} \in \mathbb{R}^{d \times d}$ for $k = 1, \dots, p$ 
    \State Sample $\mathbf{B}_j^{(c)} \in \mathbb{R}^{d \times q}$ for $j = 0, \dots, s$
    \State Partition $[\omega_{min}, \omega_{max}]$ into $C$ buckets $\{\mathcal{W}_c\}_{c=1}^C$
    \For{$c = 1, \dots, C$}
        \For{$i = 1, \dots, n$}
            \State Sample base exogenous signal $\mathbf{\tilde{x}}_t \in \mathbb{R}^q$ for $t = 1, \dots, T$
            \State Sample signal variation $\omega_i \sim \mathcal{U}(\mathcal{W}_c)$
            \State Apply signal variation $\omega_i$ to base exogenous signal $\mathbf{\tilde{x}_i}$ to obtain $\mathbf{x_t^{(i)}}$
            \State Unroll VARX using $\mathbf{A}_k^{(c)}, \mathbf{B}_j^{(c)}$ to obtain $\mathbf{y}_1, \dots, \mathbf{y}_T$
            \State Assign class label $y_i \leftarrow c$
        \EndFor
    \EndFor
\EndFor
\State \Return $\{(\mathbf{y}_i^{(1:T)}, y_i)\}$
\end{algorithmic}
\end{algorithm}

\pagebreak
\subsubsection{Relation to ARX}
The VARX process described in Section~\ref{sec:synth-data} naturally subsumes the univariate setting as a special case. For $d = 1$, the system reduces to a scalar AutoRegressive process with Exogenous inputs (ARX):
\begin{align*}
    y_t = \sum_{k=1}^{p} a_k y_{t-k} + \sum_{j=0}^{s} \mathbf{b}_j^\top \mathbf{x}_{t-j} + \varepsilon_t, \qquad \varepsilon_t \sim \mathcal{N}(0, \sigma^2)
\end{align*}
where $a_k \in \mathbb{R}$ and $\mathbf{b}_j \in \mathbb{R}^q$ are the scalar and exogenous coefficient vectors respectively. All structural properties of our original VARX prior, the class-discriminative structural variations as well as the signal variations, carry over directly to this special case without modification. This ensures that the same DGP underlies both the univariate and multivariate synthetic data, and that results on univariate benchmarks such as UCR are fully consistent with the general VARX formulation. We note, however, that the limited transfer from multivariate synthetic pre-training data to real-world multivariate classification tasks remains a limitation of \timee and identify it as an active avenue for future research.

\subsubsection{Multivariate Synthetic Data} \label{app:mv-synth}
\begin{figure}[htbp]
    \centering
      \includegraphics[width=1.0\linewidth]{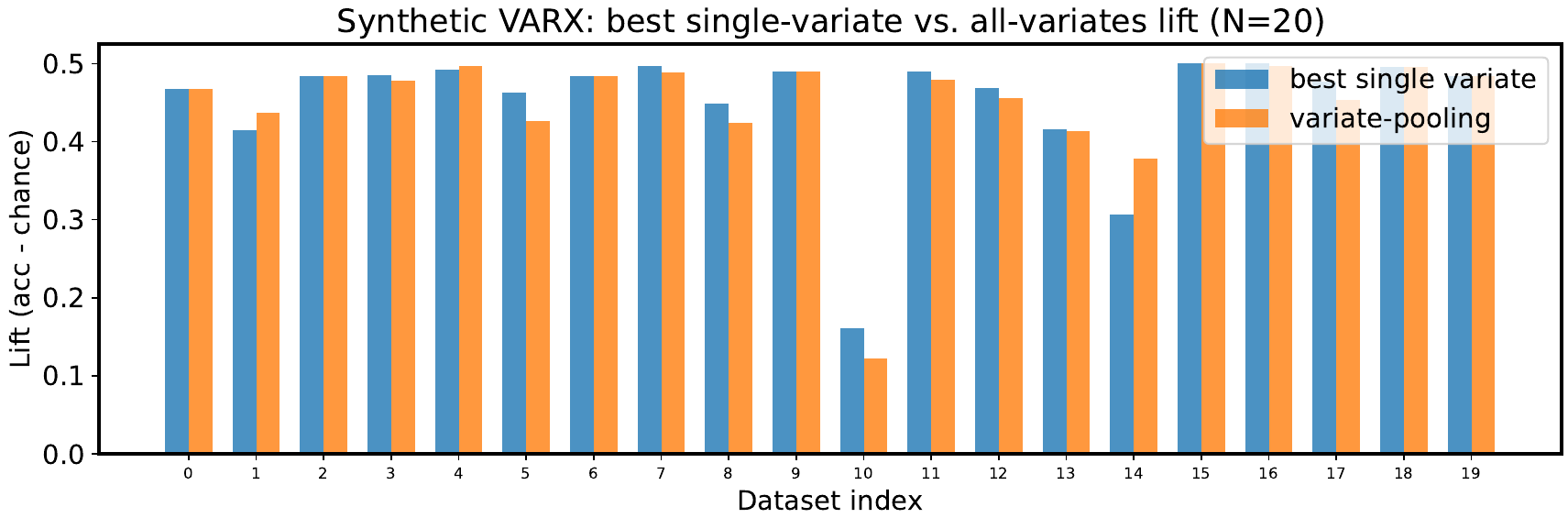}
    \caption{Dataset generated using structural variation mechanism.}
    \label{fig:synth-mv-varx}
\end{figure}

Our VARX-based prior natively supports multivariate data generation for arbitrary $d>1$, and using the extension discussed in Section~\ref{sec:abl-mv} \timee can in principle be applied to multivariate time series classification tasks directly. However, we observe that in synthetically generated VARX datasets, the information necessary for class discrimination is typically well-concentrated in individual variates already. As a consequence, averaging predictions across all $d$ variates rarely improves over the best single-variate prediction, and often even degrades performance due to the inclusion of uninformative channels. This behavior is consistent with observations in the forecasting literature \citep{ansari_chronos-2_2025} and finds motivation in Takens' embedding theorem~\citep{takens2006detecting}, which states that under mild conditions, an observation function of a dynamical system suffices to reconstruct the full state space of that system. While Takens' theorem applies strictly to deterministic smooth dynamical systems, the underlying intuition is suggestive in our setting: in the VARX process, each observed $y_t^{(i)}$ is a linear combination of the full history of all endogenous and exogenous variables, meaning inter-variable information naturally propagates into individual channels through the autoregressive dynamics. We highlight this through an empirical study on 20 synthetic multivariate datasets generated by the \timee prior, shown in Figure~\ref{fig:synth-mv-varx}. For each dataset, we compare the predictive performance of the single best univariate prediction - that is, the best performance achieved by any individual variate, against predictions obtained via \textit{variate pooling} as described in Section~\ref{sec:abl-mv}. In nearly all datasets, variate pooling fails to improve over the single best variate prediction, and in several cases actively degrades performance. This confirms that under our VARX prior, class-discriminative information is already captured by individual variates, and aggregating across channels provides little to no benefit. %
\pagebreak
\subsection{Augmentations}\label{app:augmentations}
\textbf{Smoothing.} We apply a Gaussian filter with bandwidth $\sigma \sim \mathcal{U}(\sigma_{\min}, \sigma_{\max})$, independently sampled per time series.

\textbf{Spike Injection.} The number of spikes $n_s \sim \mathcal{U}\{n_{\min}, n_{\max}\}$ is sampled per series. Spike positions are drawn uniformly from $\{1, \ldots, T\}$, and each spike amplitude is sampled independently as $s \sim \mathcal{U}(s_{\min}, s_{\max})$, giving $\mathbf{y}_t' = \mathbf{y}_t + s \cdot \mathbf{1}$ for a randomly drawn position $t$.

\textbf{Nonlinear Transformation.} A nonlinearity is sampled uniformly from $\{\tanh, \sin, \text{ReLU}, |\cdot|\}$ and applied pointwise as $\mathbf{y}_t' = g(\mathbf{y}_t)$.
\subsection{Mixup}\label{app:mixup}
To further diversify our synthetic pre-training datasets, we follow the mixup scheme used to augment forecasting data \citep{ansari_chronos-2_2025, auer2025tirexzeroshotforecastinglong}. We adapt this mechanism to the classification setting by forming convex combinations of samples drawn from within the same dataset, thereby preserving label structure while increasing intra-class diversity. Concretely, given $k$ time series samples $\tilde{\mathbf{y}}^{(1)}, \ldots, \tilde{\mathbf{y}}^{(k)}$ drawn from arbitrary classes, the mixed sample is:
\begin{align*}
    \mathbf{y}^{\text{mix}} = \sum_{i=1}^{k} \lambda_i \tilde{\mathbf{y}}^{(i)}, \qquad \lambda \sim \text{Dir}(\alpha)
\end{align*}
where $\sim \mathcal{U}\{1, K_{\max}\}$ and mixing weights $\lambda$ are drawn from a Dirichlet distribution. The full procedure is described in Algorithm~\ref{alg:mixup}.
\begin{algorithm}[h]
\caption{Time Series Mixup}
\label{alg:mixup}
\begin{algorithmic}[1]
\Require Dataset $D:= \{\mathbf{y^{(1:t)}}, y\}$, Target Length $T$

\State $D_\text{out} \gets \emptyset$
\State Sample subset of time series $\{(\mathbf{y}^{(1:t)}_i, y_i)\} \subset D$

\State \textbf{Preprocessing:}
    \State \hskip1.0em $\mathbf{y_i}^{(1:t)} \gets \text{Normalize}(\mathbf{y_i}^{(1:t)})$
    \State \hskip1.0em $\mathbf{y_i}^{(1:T)} \gets \text{Resize}(\mathbf{y_i}^{(1:t)}, T)$ \Comment{Interpolate or truncate}

\State \textbf{Mixture Generation:}
    \State \hskip1.0em Select $k$ samples $\{\mathbf{y}^{(1:T)}_j\}_{j=1}^k$ with labels $\{y_j\}_{j=1}^k$
    \State \hskip1.0em  Sample weights $\lambda \sim \text{Dirichlet}(\alpha)$
    \State \hskip1.0em $\mathbf{y}^{(1:T)}_{mix} \gets \sum_{j=1}^k (\lambda_j \cdot \mathbf{y}^{(1:T)}_j) + \mathcal{N}(\mathbf{0}, \mathbf{\Sigma})$

\State \textbf{Pseudo-Label Assignment:}
\State \hskip1.0em Assign $y_{mix}$ based on $\lambda$ and $\{y_j\}$ using threshold

\State \hskip1.0em Append $(\mathbf{y}^{(1:T)}_{mix}, y_{mix})$ to $D_{\text{out}}$

\State \textbf{return} $D_\text{out}$
\end{algorithmic}
\end{algorithm}

\subsection{Examples of synthetic data}
\begin{figure}[htbp]
    \centering
    \includegraphics[width=1.0\linewidth]{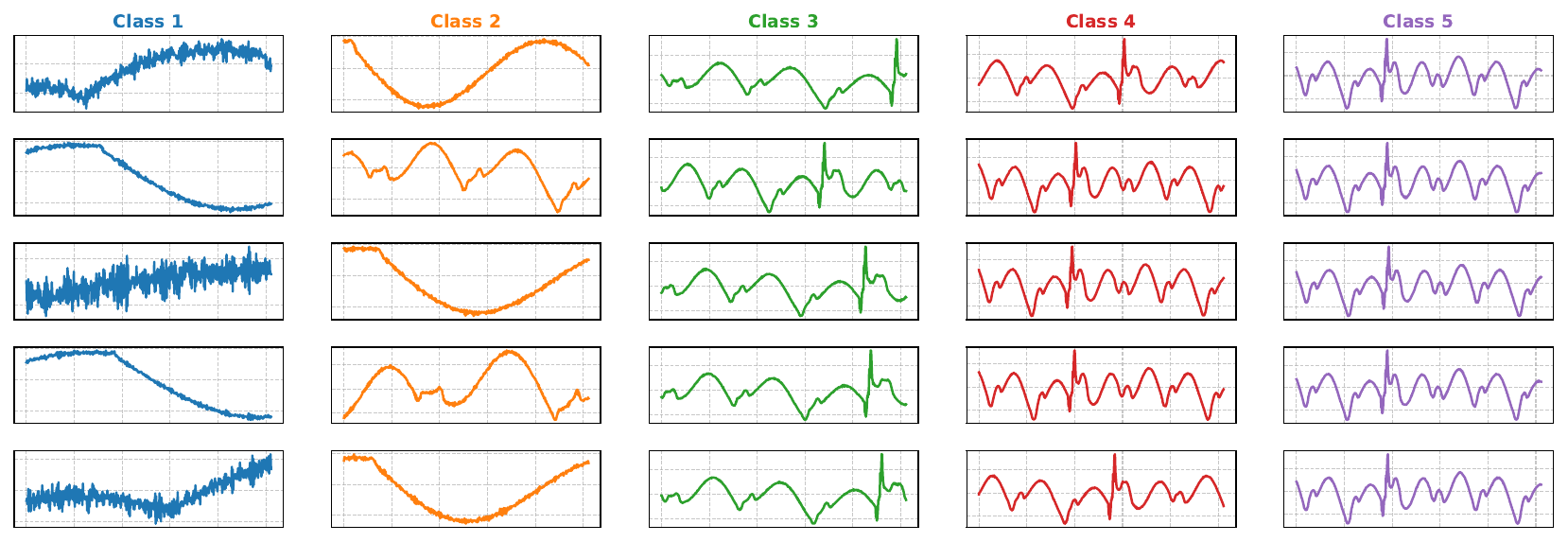}
    \caption{Dataset generated using structural variation mechanism.}
    \label{fig:int-b-1}
\end{figure}

\begin{figure}[htbp]
    \centering
    \includegraphics[width=1.0\linewidth]{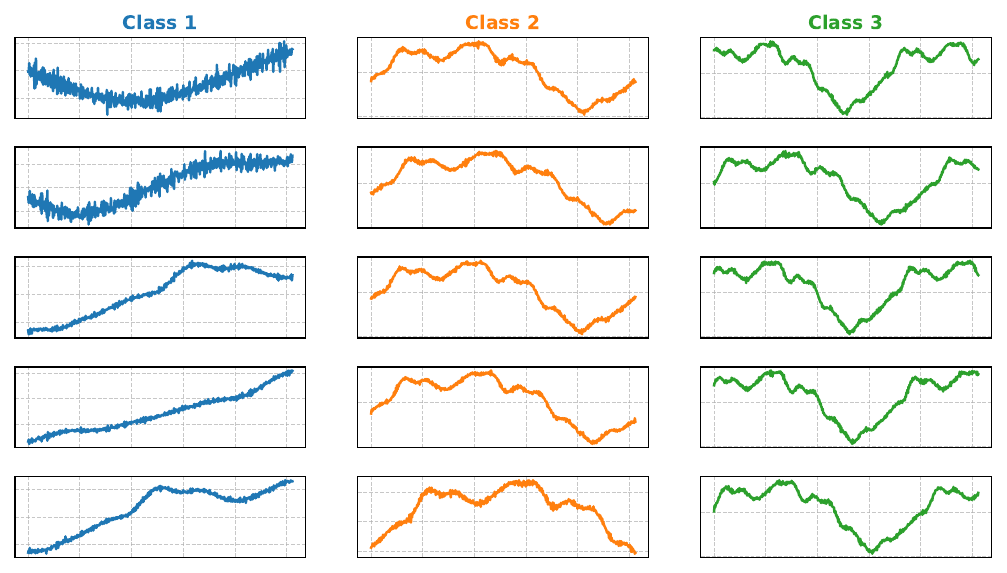}
    \caption{Dataset generated using structural variation mechanism.}
    \label{fig:int-b-2}
\end{figure}

\begin{figure}[htbp]
    \centering
    \includegraphics[width=0.85\linewidth]{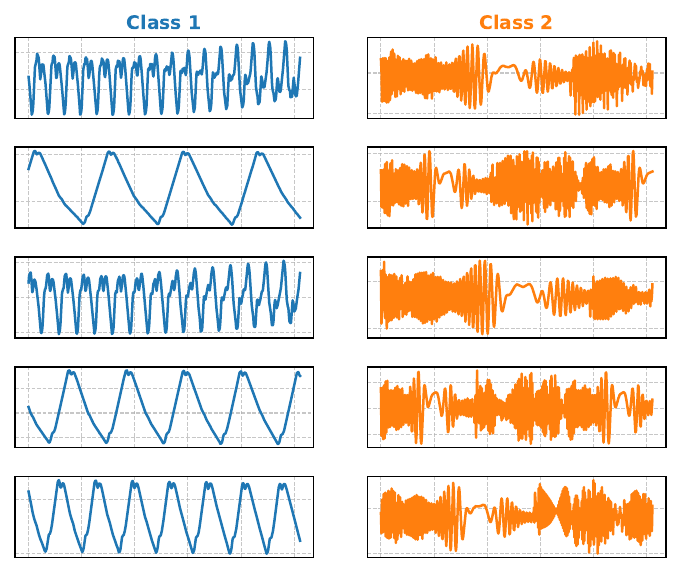}
    \caption{Dataset generated using signal variation mechanism.}
    \label{fig:warp}
\end{figure}

\begin{figure}[htbp]
    \centering
    \includegraphics[width=0.85\linewidth]{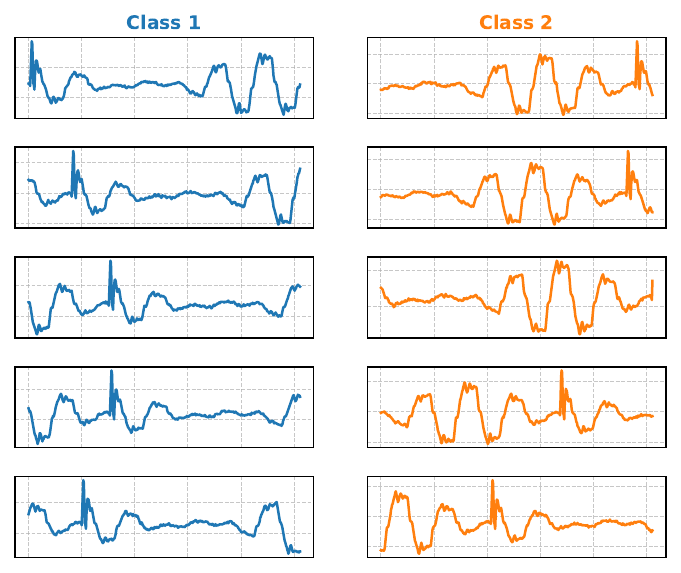}
    \caption{Dataset generated using signal variation mechanism.}
    \label{fig:shift}
\end{figure}

\begin{figure}[htbp]
     \centering
     \begin{subfigure}[b]{0.48\textwidth}
         \centering
         \includegraphics[width=1.0\linewidth]{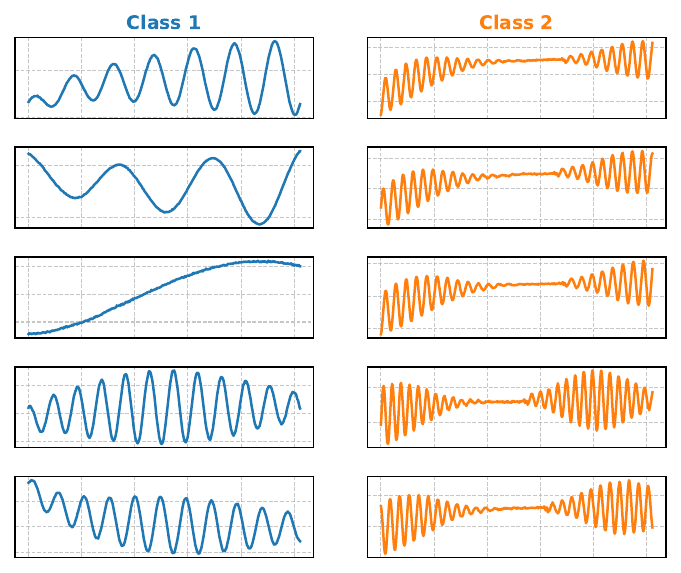}
         \caption{Dataset generated using signal variation before applying the mixup scheme.}
         \label{fig:no-mixup}
     \end{subfigure}
     \hfill %
     \begin{subfigure}[b]{0.48\textwidth}
         \centering
         \includegraphics[width=1.0\linewidth]{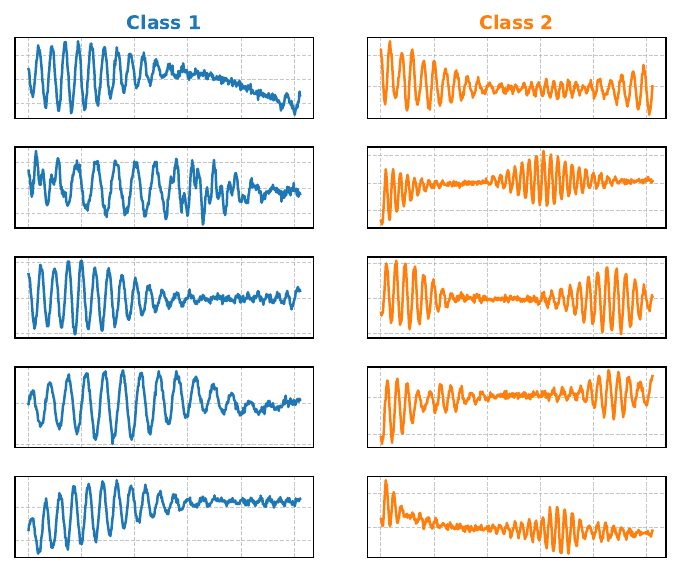}
         \caption{Dataset generated using signal variation after applying the mixup scheme.}
         \label{fig:mixup}
     \end{subfigure}
\end{figure}

We present example datasets generated from our synthetic prior, for different class generation mechanisms. Figures~\ref{fig:int-b-1},~\ref{fig:int-b-2} present samples generated by structural variation. Figures~\ref{fig:warp}~\ref{fig:shift} present datasets generated by signal level variation. Further we showcase the impact of the mixup scheme described in Section \ref{app:mixup} on a generated synthetic dataset, where Figure~\ref{fig:no-mixup} shows the originally sampled dataset before applying mixup while Figure~\ref{fig:mixup} shows the same dataset after applying mixup.


\newpage
\section{Pretraining and Inference}
\subsection{Pretraining Configurations} \label{app:pretraining}

Table~\ref{tab:hyperparams} summarizes the full set of hyperparameters used during pretraining.

\begin{table}[h]
\centering
\caption{Pretraining hyperparameters.}
\label{tab:hyperparams}
\begin{tabular}{ll}
\toprule
\textbf{Hyperparameter} & \textbf{Value} \\
\midrule
Training steps          & 30{,}000 \\
Warmup steps            & 1{,}500 (5\% of total) \\
Learning rate           & $3 \times 10^{-4}$ \\
LR schedule             & Cosine decay with linear warmup \\
Optimizer (matrix weights)     & Muon~\citep{jordan2024muon} \\
Optimizer (non-matrix weights) & AdamW~\citep{loshchilov2019decoupledweightdecayregularization} \\
Batch size              & 240 \\
Mixed precision         & bf16 AMP \\
Number of classes $C$   & Uniform $\sim \mathcal{U}\{2, 10\}$ per task \\
Support-query split     & Uniform $\sim \mathcal{U}[0.1, 0.9]$ per task \\
Series length           & Uniform $\sim \mathcal{U}[16, 1024]$, shared across context \\
Hardware                & Single NVIDIA H200 GPU \\
Training time           & $\approx 40$ hours \\
\bottomrule
\end{tabular}

\end{table}

\subsection{Inference Pipeline}\label{app:inference-pipeline}
At inference, we found it beneficial to apply lightweight preprocessing transformations and ensemble predictions across differently preprocessed inputs. Concretely, we consider two elementary transformations: interpolation of short time series to a target length in \texttt{\{128,256,512\}}, and first-order differencing. %
Each ensemble member is formed from one of the individual transformation (or compositions of multiple individuals). Final predictions are obtained by averaging the predicted distributions across all members. By default, \timee ensembles 4 preprocessing variants.

\newpage

\section{Implementation of Baselines}\label{app:baselines}
\subsection{Statistical Methods}                                                                          
  \textbf{MiniRocket}~\citep{dempster2021minirocket}: We apply MiniRocket feature extraction followed by a
   \texttt{StandardScaler} and \texttt{RidgeClassifierCV}, using the \texttt{aeon} implementation. \par     
  \textbf{Hydra}~\citep{dempster2023hydra}: We use the \texttt{aeon} implementation with default          
  hyperparameters. \par                                                                                     
  \textbf{Catch22}~\citep{lubba2019catch22canonicaltimeseriescharacteristics}: We extract 22 canonical time-series summary statistics via 
  \texttt{pycatch22} and classify with a \texttt{RandomForestClassifier} (\texttt{n\_estimators=300}). \par 
  
  \textbf{DTW-1NN}~\citep{sakoe1978dynamic}: 1-nearest-neighbour classifier with Dynamic Time Warping distance, using the \texttt{aeon} implementation.                                                         
\subsection{Deep Learning Methods}
  \textbf{TS2Vec}~\citep{yue2022ts2vecuniversalrepresentationtime}: A contrastive self-supervised encoder trained independently per dataset with output dimension 320. We extract full-series embeddings and classify with an SVM whose regularisation parameter $C$ is selected via 5-fold cross-validation on the training set. \par
  \textbf{InceptionTime}~\citep{Ismail_Fawaz_2020}: An ensemble of 5 InceptionTime networks trained for 1500 epochs per dataset, using the \texttt{aeon} implementation.   

\subsection{Foundation Models}
For all foundation models we use a \texttt{RandomForestClassifier} from \texttt{scikit-learn}~\citep{scikit-learn} with \texttt{n\_estimators=300} consistent with~\citet{feofanov2026mantisv2closingzeroshotgap}. All remaining hyperparameters take their default values.
Where a tuned variant is reported, we select the encoder layer whose embeddings yield the highest mean ROCAUC across the evaluation suite and apply it uniformly across all datasets. 

\subsubsection{TSC Foundation Models}
\textbf{Mantis-V2}~\citep{feofanov2026mantisv2closingzeroshotgap}: We use the official implementation.~\footnote{\url{https://github.com/vfeofanov/mantis}} As Mantis-V2 requires input sequences of length divisible by \texttt{num\_patches}, we interpolate to next higher num patches. For the tuned version, embeddings from the 3rd layer yielded the strongest downstream performance. We also interpolate each time series to a minimum length of 512. \par

\textbf{Moment}~\citep{goswami_moment_2024}: We use the \textit{large} size model for all experiments with
   Moment. We follow all defaults from the official
  implementation.~\footnote{\url{https://github.com/moment-timeseries-foundation-model/moment}} Using later       
  encoder layers performs comparably; we use the 13th layer for the univariate setting and the 11th for the
  multivariate setting. \par

\textbf{NuTime}~\citep{lin2024nutimenumericallymultiscaledembedding}: A window-based ViT pretrained via BYOL on UCR datasets. We      
  extract the frozen CLS-token embedding and classify with the above classifier. Input series are resampled 
  to length 512. For the tuned variant, embeddings from the 6th layer yielded the strongest downstream
  performance. \par                                                                                         
  \textbf{TiCT}~\citep{yeh2025tictsyntheticallypretrainedfoundation}: A 47M-parameter ResNet pretrained for in-context time series
  classification. For each test sample, we retrieve its 64 nearest neighbours from the training set as      
  context and perform a single forward pass; no separate classifier is required.

\subsubsection{Forecasting Models}
For both forecasting models, we extract patch-level embeddings per time series, apply mean pooling over patches to obtain a fixed-size representation, and classify using the above mentioned classifier.\par
\textbf{Chronos-2}~\citep{ansari_chronos-2_2025}: We use the \textit{base} model throughout. By default, embeddings are extracted from the final encoder layer. For the tuned variant, we take embeddings from the 5th layer, which yielded the strongest downstream performance. \par
\textbf{TiRex}~\citep{auer2025tirexzeroshotforecastinglong}: We apply the same extraction and classification strategy. For the tuned variant, we likewise use the 6th layer, following the same selection procedure.

\subsection{Multivariate Setting}                                                                         
For models where we implement the multivariate adaptation ourselves, we reuse the best layer settings identified in the univariate search. Per-channel embeddings are concatenated before classification; mean pooling consistently degraded performance across models. We use a \texttt{RandomForestClassifier} throughout, as SVMs struggled with the extended embedding dimension produced by concatenation. \textbf{TiCT} classifies each channel independently and averages the per-channel softmax probabilities. DTW-1NN, MiniRocket, Hydra, and InceptionTime use their native multivariate implementations via \texttt{aeon} without modification. For TiRex, reusing the univariate tuned settings did not transfer to the multivariate setting and the default outperforms the tuned version.

\section{Experiment Results}\label{app:experiment-results}
\subsection{Inference Time vs. Performance on datasets}\label{app:inference-time}

To evaluate the trade-off between the inference efficiency and the predictive performance, we measure the inference time of each method.

For GPU baselines, we evaluate all GPU-accelerated candidates on a NVIDIA H100 GPU; for CPU baselines, we evaluate them on 32-cores AMD EPYC 9655.

\begin{figure}[h]
    \centering
    \includegraphics[width=0.8\linewidth]{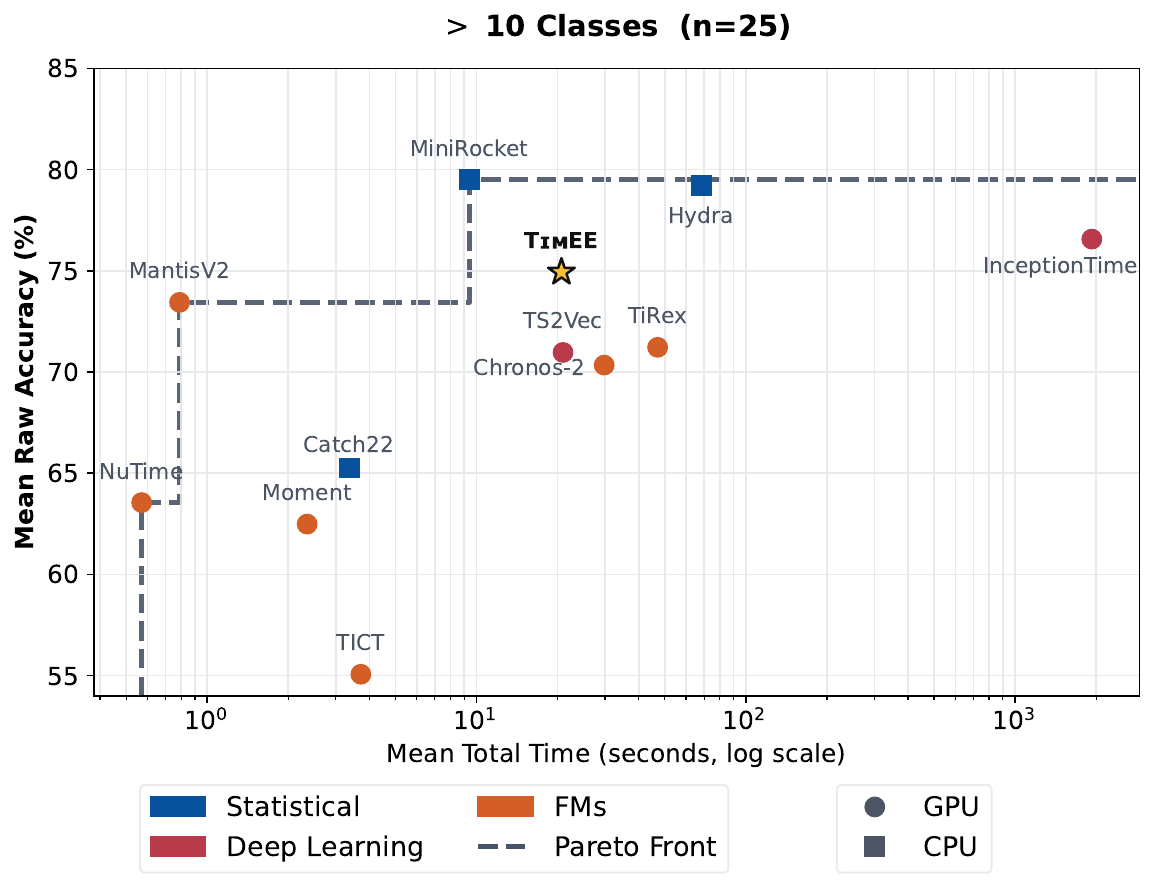}
    \caption{Pareto front of inference speed vs. predictive performance on UCR datasets with more than 10 classes.
    \label{fig:pareto-front-more-than-10}
    OneVsRest many-class strategy of \timee imposes significant inference overhead.
    This remains as a limitation of \timee.
}
    \label{fig:pareto-large}
\end{figure}

\pagebreak
\subsection{Critical Difference Diagram}\label{app:cd-diagram}                                            
  \begin{figure}[h]                                
      \centering  
      \includegraphics[width=0.85\linewidth]{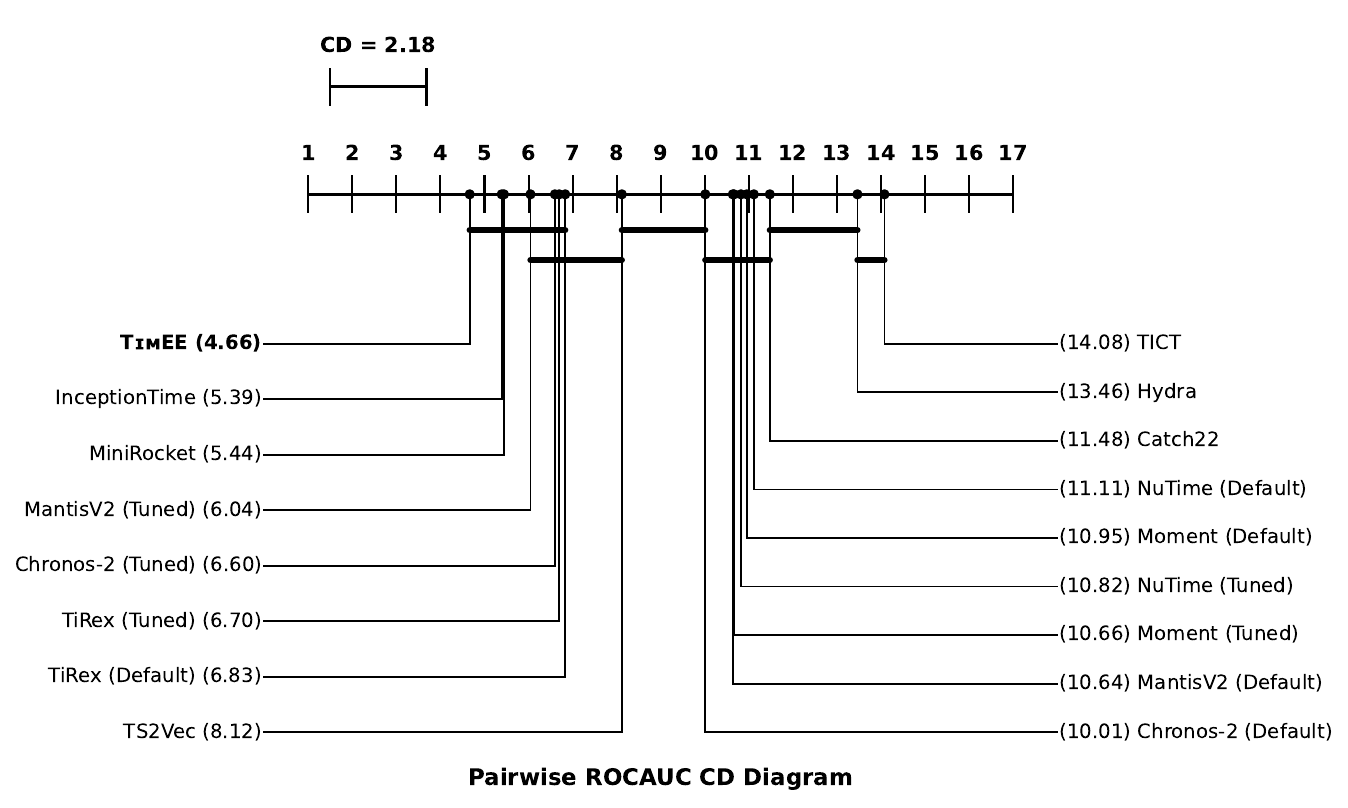}
      \caption{Critical difference diagram over ROC AUC across 128 UCR datasets (Wilcoxon--Holm post-hoc test, $\alpha = 0.05$). Methods connected by a  horizontal bar are not significantly different.}      \label{fig:cd-diagram-roc}             
  \end{figure} 

  \begin{figure}[h]
    \centering  
      \includegraphics[width=0.85\linewidth]{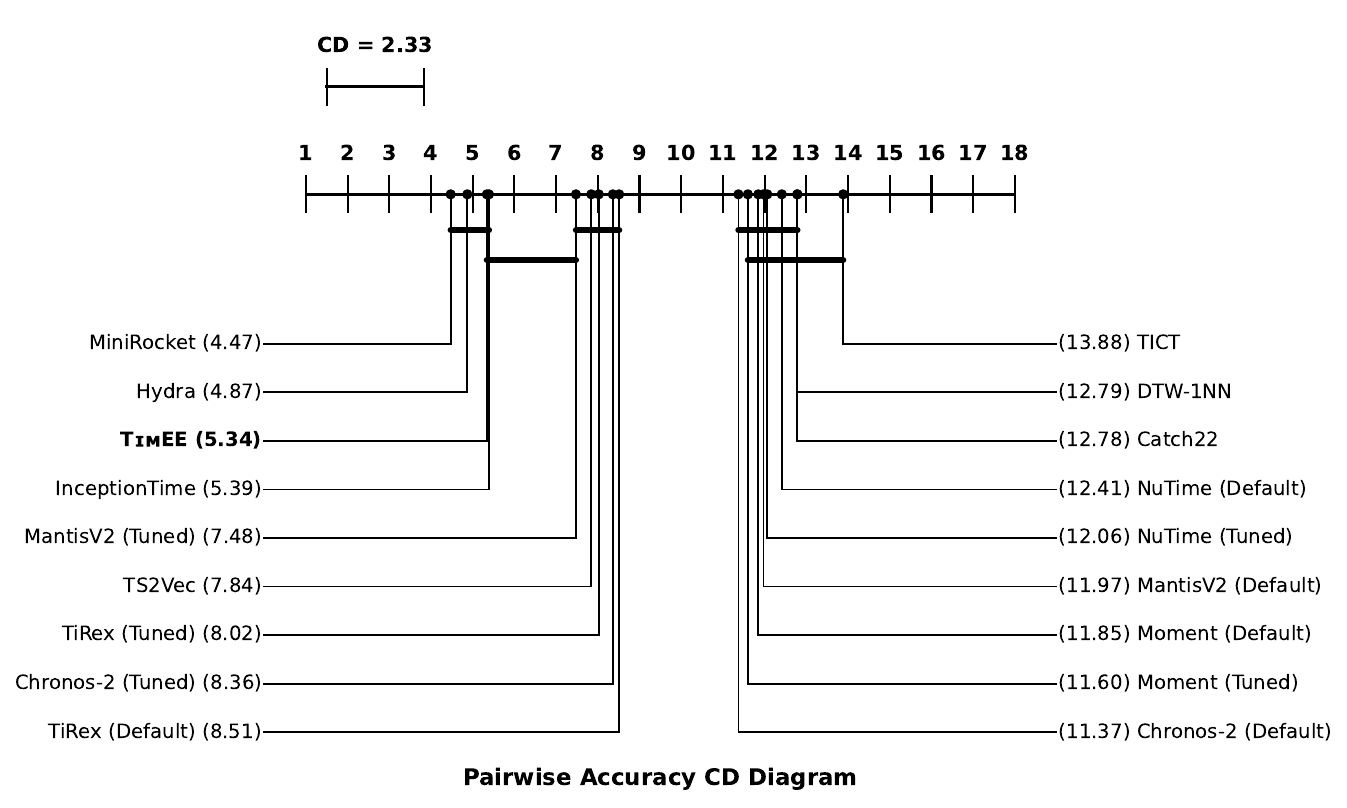}
      \caption{Critical difference diagram over ROC AUC across 128 UCR datasets (Wilcoxon--Holm post-hoc test, $\alpha = 0.05$). Methods connected by a  horizontal bar are not significantly different.}  
      \label{fig:cd-diagram-accuracy}    
   \end{figure}

\subsection{Per-dataset results}\label{app:per-dataset-res}

\begin{table}[htbp]
\centering
\setlength{\tabcolsep}{4pt}
\caption{UCR univariate ROCAUC. \textbf{Bold} = best per dataset.}
\label{tab:ucr-rocauc}
\resizebox{\linewidth}{!}{%
\begin{tabular}{@{}lccccccccccccc@{}}
\toprule
Dataset & DTW-1NN & Catch22 & MiniRocket & Hydra & InceptionTime & TS2Vec & TICT & Chronos-2 & TiRex & Moment & NuTime & MantisV2 & TimEE \\
\midrule
ACSF1 & -- & 0.986 & 0.996 & 0.917 & \textbf{0.996} & 0.989 & 0.777 & 0.989 & 0.990 & 0.976 & 0.972 & 0.977 & 0.983 \\
Adiac & -- & 0.985 & 0.979 & 0.911 & 0.992 & 0.981 & 0.922 & 0.994 & 0.992 & 0.989 & 0.988 & \textbf{0.994} & 0.992 \\
AllGestureWiimoteX & -- & 0.886 & 0.940 & 0.835 & \textbf{0.962} & 0.949 & 0.876 & 0.912 & 0.922 & 0.902 & 0.907 & 0.945 & 0.947 \\
AllGestureWiimoteY & -- & 0.901 & 0.959 & 0.851 & \textbf{0.976} & 0.946 & 0.874 & 0.922 & 0.944 & 0.924 & 0.883 & 0.947 & 0.960 \\
AllGestureWiimoteZ & -- & 0.899 & 0.951 & 0.825 & \textbf{0.972} & 0.954 & 0.854 & 0.934 & 0.935 & 0.917 & 0.911 & 0.956 & 0.954 \\
ArrowHead & -- & 0.892 & \textbf{0.978} & 0.883 & 0.976 & 0.937 & 0.895 & 0.936 & 0.964 & 0.945 & 0.912 & 0.940 & 0.975 \\
BME & -- & 0.983 & \textbf{1.000} & \textbf{1.000} & \textbf{1.000} & 1.000 & 0.848 & \textbf{1.000} & \textbf{1.000} & 0.991 & 0.960 & \textbf{1.000} & \textbf{1.000} \\
Beef & -- & 0.801 & 0.954 & 0.896 & 0.942 & 0.889 & 0.858 & 0.928 & 0.992 & 0.924 & 0.942 & 0.915 & \textbf{0.997} \\
BeetleFly & -- & 0.850 & \textbf{1.000} & 0.900 & \textbf{1.000} & 0.990 & 0.880 & 0.970 & \textbf{1.000} & \textbf{1.000} & 0.950 & \textbf{1.000} & \textbf{1.000} \\
BirdChicken & -- & 0.980 & 0.900 & 0.900 & \textbf{1.000} & 0.800 & 0.990 & 0.970 & 0.940 & 0.980 & \textbf{1.000} & 0.990 & 0.960 \\
CBF & -- & 0.997 & 1.000 & 0.995 & \textbf{1.000} & \textbf{1.000} & 0.992 & 1.000 & 1.000 & 0.996 & 0.999 & 1.000 & 1.000 \\
Car & -- & 0.904 & \textbf{0.968} & 0.953 & 0.955 & 0.909 & 0.836 & 0.934 & 0.940 & 0.911 & 0.932 & 0.941 & 0.956 \\
Chinatown & -- & 0.987 & 0.995 & 0.988 & 0.996 & 0.989 & 0.977 & \textbf{0.996} & 0.995 & 0.995 & 0.982 & 0.989 & 0.995 \\
ChlorineConcentration & -- & 0.677 & 0.878 & 0.785 & \textbf{0.947} & 0.919 & 0.711 & 0.851 & 0.874 & 0.832 & 0.789 & 0.831 & 0.803 \\
CinCECGTorso & -- & 0.949 & 0.970 & 0.997 & 0.971 & 0.950 & 0.862 & 1.000 & 1.000 & 0.900 & 0.947 & 0.959 & \textbf{1.000} \\
Coffee & -- & \textbf{1.000} & \textbf{1.000} & \textbf{1.000} & \textbf{1.000} & \textbf{1.000} & 0.995 & \textbf{1.000} & \textbf{1.000} & \textbf{1.000} & 0.995 & \textbf{1.000} & \textbf{1.000} \\
Computers & -- & 0.806 & 0.794 & 0.716 & \textbf{0.880} & 0.744 & 0.744 & 0.828 & 0.808 & 0.733 & 0.832 & 0.794 & 0.722 \\
CricketX & -- & 0.906 & 0.974 & 0.895 & \textbf{0.984} & 0.976 & 0.894 & 0.960 & 0.953 & 0.954 & 0.946 & 0.977 & 0.981 \\
CricketY & -- & 0.928 & 0.980 & 0.914 & \textbf{0.989} & 0.971 & 0.905 & 0.966 & 0.967 & 0.960 & 0.959 & 0.978 & 0.988 \\
CricketZ & -- & 0.914 & 0.974 & 0.889 & \textbf{0.988} & 0.973 & 0.917 & 0.958 & 0.957 & 0.956 & 0.950 & 0.981 & 0.986 \\
Crop & -- & 0.961 & 0.973 & 0.861 & 0.981 & 0.980 & 0.942 & 0.977 & 0.980 & 0.972 & 0.970 & 0.979 & \textbf{0.984} \\
DiatomSizeReduction & -- & 0.995 & 0.975 & 0.952 & 0.997 & 1.000 & 0.969 & 0.999 & 0.991 & 0.997 & 0.993 & 0.986 & \textbf{1.000} \\
DistalPhalanxOutlineAgeGroup & -- & 0.865 & 0.892 & 0.830 & 0.858 & 0.881 & 0.856 & 0.904 & 0.899 & 0.881 & 0.877 & \textbf{0.906} & 0.886 \\
DistalPhalanxOutlineCorrect & -- & 0.871 & 0.871 & 0.768 & 0.851 & 0.787 & 0.825 & \textbf{0.878} & 0.866 & 0.877 & 0.842 & 0.869 & 0.876 \\
DistalPhalanxTW & -- & 0.845 & 0.879 & 0.741 & 0.854 & 0.866 & 0.818 & 0.874 & 0.875 & 0.878 & 0.858 & 0.885 & \textbf{0.891} \\
DodgerLoopDay & -- & 0.829 & 0.903 & 0.717 & 0.852 & 0.862 & 0.850 & 0.879 & 0.905 & 0.808 & 0.819 & 0.863 & \textbf{0.923} \\
DodgerLoopGame & -- & 0.765 & 0.941 & 0.817 & 0.917 & \textbf{0.980} & 0.916 & 0.942 & 0.921 & 0.880 & 0.848 & 0.886 & 0.972 \\
DodgerLoopWeekend & -- & \textbf{0.997} & 0.984 & 0.984 & 0.992 & 0.991 & 0.991 & 0.997 & 0.992 & 0.988 & 0.985 & 0.991 & 0.991 \\
ECG200 & -- & 0.908 & 0.965 & 0.850 & \textbf{0.967} & 0.957 & 0.882 & 0.927 & 0.934 & 0.946 & 0.905 & 0.928 & 0.957 \\
ECG5000 & -- & 0.911 & 0.886 & 0.762 & 0.927 & 0.936 & 0.884 & 0.942 & \textbf{0.953} & 0.936 & 0.911 & 0.916 & 0.952 \\
ECGFiveDays & -- & 0.956 & \textbf{1.000} & \textbf{1.000} & \textbf{1.000} & \textbf{1.000} & 0.921 & 0.996 & 1.000 & 0.967 & 0.914 & 0.984 & 1.000 \\
EOGHorizontalSignal & -- & 0.905 & 0.894 & 0.767 & 0.930 & 0.903 & 0.805 & 0.909 & 0.915 & 0.886 & 0.865 & \textbf{0.937} & 0.929 \\
EOGVerticalSignal & -- & 0.870 & 0.835 & 0.724 & 0.879 & 0.862 & 0.787 & 0.871 & \textbf{0.907} & 0.831 & 0.778 & 0.857 & 0.891 \\
Earthquakes & -- & 0.692 & 0.672 & 0.490 & 0.662 & 0.544 & 0.637 & 0.673 & 0.719 & 0.635 & 0.709 & 0.613 & \textbf{0.730} \\
ElectricDevices & -- & 0.911 & 0.890 & 0.809 & 0.894 & 0.916 & 0.840 & \textbf{0.918} & 0.902 & 0.893 & 0.900 & 0.912 & 0.897 \\
EthanolLevel & -- & 0.617 & 0.811 & 0.726 & \textbf{0.958} & 0.695 & 0.597 & 0.686 & 0.852 & 0.638 & 0.631 & 0.614 & 0.878 \\
FaceAll & -- & 0.966 & 0.997 & 0.968 & 0.995 & 0.994 & 0.966 & 0.988 & 0.997 & 0.972 & 0.949 & 0.963 & \textbf{0.999} \\
FaceFour & -- & 0.942 & \textbf{1.000} & 0.931 & 0.992 & 0.972 & 0.940 & 0.982 & \textbf{1.000} & 0.964 & 0.989 & 0.993 & \textbf{1.000} \\
FacesUCR & -- & 0.949 & 0.998 & 0.966 & \textbf{0.999} & 0.995 & 0.931 & 0.976 & 0.989 & 0.958 & 0.958 & 0.982 & 0.995 \\
FiftyWords & -- & 0.934 & 0.983 & 0.861 & \textbf{0.991} & 0.958 & 0.905 & 0.973 & 0.965 & 0.972 & 0.953 & 0.971 & 0.985 \\
Fish & -- & 0.965 & 0.998 & 0.994 & \textbf{1.000} & 0.992 & 0.944 & 0.988 & 0.994 & 0.989 & 0.997 & 0.991 & 0.996 \\
FordA & -- & 0.967 & 0.988 & 0.961 & \textbf{0.992} & 0.983 & 0.958 & 0.977 & 0.991 & 0.962 & 0.962 & 0.981 & 0.987 \\
FordB & -- & 0.845 & 0.891 & 0.825 & \textbf{0.937} & 0.875 & 0.807 & 0.872 & 0.924 & 0.846 & 0.842 & 0.881 & 0.901 \\
FreezerRegularTrain & -- & 1.000 & 1.000 & 0.998 & 1.000 & 0.998 & 0.971 & 0.999 & \textbf{1.000} & 0.968 & 0.999 & 1.000 & 1.000 \\
FreezerSmallTrain & -- & \textbf{0.995} & 0.995 & 0.926 & 0.900 & 0.956 & 0.851 & 0.973 & 0.983 & 0.861 & 0.995 & 0.987 & 0.991 \\
Fungi & -- & 0.997 & \textbf{1.000} & \textbf{1.000} & \textbf{1.000} & 0.008 & 0.988 & 0.996 & 0.997 & 0.999 & 0.974 & 0.997 & 1.000 \\
GestureMidAirD1 & -- & \textbf{0.981} & 0.975 & 0.872 & 0.958 & 0.920 & 0.958 & 0.981 & 0.978 & 0.972 & 0.947 & 0.974 & 0.980 \\
GestureMidAirD2 & -- & 0.978 & 0.970 & 0.836 & 0.938 & 0.880 & 0.946 & 0.983 & \textbf{0.984} & 0.982 & 0.955 & 0.979 & 0.982 \\
GestureMidAirD3 & -- & 0.928 & 0.898 & 0.732 & 0.807 & 0.740 & 0.827 & \textbf{0.959} & 0.951 & 0.917 & 0.873 & 0.938 & 0.938 \\
GesturePebbleZ1 & -- & 0.968 & 0.989 & 0.933 & 0.987 & 0.949 & 0.962 & 0.989 & 0.989 & 0.981 & 0.961 & \textbf{0.990} & 0.979 \\
GesturePebbleZ2 & -- & 0.958 & 0.983 & 0.929 & 0.970 & 0.965 & 0.957 & 0.990 & 0.992 & 0.981 & 0.971 & \textbf{0.993} & 0.970 \\
GunPoint & -- & 0.994 & \textbf{1.000} & \textbf{1.000} & \textbf{1.000} & 0.998 & 0.985 & \textbf{1.000} & 0.999 & 0.997 & 0.994 & 0.998 & 0.996 \\
GunPointAgeSpan & -- & 0.995 & 1.000 & \textbf{1.000} & 1.000 & 0.999 & 0.999 & 1.000 & 0.998 & 0.992 & 0.997 & \textbf{1.000} & 1.000 \\
GunPointMaleVersusFemale & -- & 1.000 & \textbf{1.000} & \textbf{1.000} & \textbf{1.000} & \textbf{1.000} & 1.000 & 1.000 & 1.000 & 0.999 & 0.998 & \textbf{1.000} & \textbf{1.000} \\
GunPointOldVersusYoung & -- & 0.997 & \textbf{1.000} & \textbf{1.000} & \textbf{1.000} & \textbf{1.000} & 0.996 & \textbf{1.000} & \textbf{1.000} & 0.995 & \textbf{1.000} & \textbf{1.000} & \textbf{1.000} \\
Ham & -- & 0.672 & 0.812 & 0.742 & \textbf{0.828} & 0.770 & 0.718 & 0.783 & 0.825 & 0.792 & 0.825 & 0.733 & 0.780 \\
HandOutlines & -- & 0.918 & 0.961 & 0.927 & \textbf{0.983} & 0.948 & 0.929 & 0.953 & 0.947 & 0.945 & 0.939 & 0.952 & 0.966 \\
Haptics & -- & 0.746 & 0.813 & 0.695 & \textbf{0.821} & 0.792 & 0.733 & 0.807 & 0.797 & 0.776 & 0.775 & 0.806 & 0.817 \\
Herring & -- & 0.511 & \textbf{0.768} & 0.735 & 0.734 & 0.641 & 0.654 & 0.740 & 0.720 & 0.659 & 0.656 & 0.766 & 0.526 \\
HouseTwenty & -- & 0.983 & 0.991 & 0.956 & 0.972 & 0.957 & 0.989 & 0.982 & 0.991 & 0.975 & 0.956 & \textbf{0.992} & 0.988 \\
InlineSkate & -- & 0.808 & 0.814 & 0.703 & 0.841 & 0.781 & 0.738 & 0.804 & \textbf{0.843} & 0.712 & 0.742 & 0.821 & 0.802 \\
InsectEPGRegularTrain & -- & 0.997 & \textbf{1.000} & \textbf{1.000} & \textbf{1.000} & \textbf{1.000} & 0.983 & \textbf{1.000} & \textbf{1.000} & 0.980 & \textbf{1.000} & \textbf{1.000} & 0.998 \\
InsectEPGSmallTrain & -- & 0.959 & \textbf{1.000} & \textbf{1.000} & \textbf{1.000} & \textbf{1.000} & 0.899 & 0.988 & 0.998 & 0.972 & \textbf{1.000} & \textbf{1.000} & 0.992 \\
InsectWingbeatSound & -- & 0.930 & 0.947 & 0.810 & 0.938 & 0.949 & 0.866 & 0.953 & 0.959 & 0.946 & 0.900 & 0.942 & \textbf{0.959} \\
ItalyPowerDemand & -- & 0.959 & 0.993 & 0.966 & 0.990 & 0.985 & 0.934 & 0.993 & 0.993 & 0.990 & 0.950 & 0.982 & \textbf{0.994} \\
LargeKitchenAppliances & -- & 0.953 & 0.957 & 0.912 & \textbf{0.979} & 0.951 & 0.853 & 0.972 & 0.932 & 0.900 & 0.891 & 0.925 & 0.867 \\
Lightning2 & -- & 0.869 & 0.803 & 0.715 & 0.905 & \textbf{0.937} & 0.789 & 0.824 & 0.815 & 0.861 & 0.778 & 0.847 & 0.898 \\
Lightning7 & -- & 0.935 & 0.974 & 0.895 & 0.967 & 0.976 & 0.909 & 0.972 & 0.971 & 0.955 & 0.956 & 0.966 & \textbf{0.979} \\
Mallat & -- & 0.997 & 0.998 & 0.966 & 0.999 & 0.989 & 0.871 & 0.998 & \textbf{1.000} & 0.998 & 0.992 & 0.997 & 0.999 \\
Meat & -- & 0.974 & \textbf{1.000} & 0.938 & 0.997 & 0.998 & 0.805 & \textbf{1.000} & 0.994 & \textbf{1.000} & 0.996 & 1.000 & \textbf{1.000} \\
MedicalImages & -- & 0.964 & 0.969 & 0.837 & 0.972 & \textbf{0.974} & 0.943 & 0.962 & 0.972 & 0.963 & 0.955 & 0.968 & 0.972 \\
MelbournePedestrian & -- & 0.975 & 0.997 & 0.961 & \textbf{0.999} & 0.998 & 0.972 & 0.998 & 0.998 & 0.987 & 0.996 & 0.998 & 0.997 \\
MiddlePhalanxOutlineAgeGroup & -- & 0.650 & 0.599 & 0.601 & 0.599 & 0.649 & 0.633 & 0.645 & 0.649 & 0.627 & \textbf{0.666} & 0.645 & 0.642 \\
MiddlePhalanxOutlineCorrect & -- & 0.854 & 0.919 & 0.826 & 0.905 & 0.901 & 0.842 & \textbf{0.939} & 0.921 & 0.921 & 0.867 & 0.931 & 0.928 \\
MiddlePhalanxTW & -- & 0.753 & 0.781 & 0.639 & 0.759 & 0.768 & 0.741 & 0.783 & 0.763 & \textbf{0.790} & 0.738 & 0.779 & 0.762 \\
MixedShapesRegularTrain & -- & 0.989 & 0.997 & 0.988 & 0.998 & 0.989 & 0.986 & 0.995 & 0.998 & 0.983 & 0.992 & 0.996 & \textbf{0.999} \\
MixedShapesSmallTrain & -- & 0.975 & 0.994 & 0.977 & 0.989 & 0.978 & 0.948 & 0.989 & 0.994 & 0.962 & 0.986 & 0.990 & \textbf{0.997} \\
MoteStrain & -- & 0.946 & 0.976 & 0.927 & 0.955 & 0.933 & 0.956 & 0.981 & 0.979 & 0.971 & \textbf{0.991} & 0.987 & 0.973 \\
NonInvasiveFetalECGThorax1 & -- & 0.996 & 0.998 & 0.970 & \textbf{0.999} & 0.998 & 0.984 & 0.996 & 0.998 & 0.997 & 0.991 & 0.996 & 0.999 \\
NonInvasiveFetalECGThorax2 & -- & 0.996 & 0.998 & 0.977 & 0.999 & 0.998 & 0.988 & 0.997 & 0.998 & 0.997 & 0.994 & 0.997 & \textbf{0.999} \\
OSULeaf & -- & 0.945 & 0.996 & 0.990 & \textbf{0.997} & 0.973 & 0.925 & 0.991 & 0.995 & 0.964 & 0.960 & 0.996 & 0.990 \\
OliveOil & -- & 0.917 & 0.989 & 0.945 & 0.967 & 0.903 & 0.696 & 0.991 & 0.982 & 0.973 & 0.958 & \textbf{0.991} & 0.982 \\
\bottomrule
\end{tabular}%
}
\end{table}

\begin{table}[htbp]
\centering
\setlength{\tabcolsep}{4pt}
\caption{UCR univariate ROCAUC. \textbf{Bold} = best per dataset. (continued)}
\label{tab:ucr-rocauc-cont}
\resizebox{\linewidth}{!}{%
\begin{tabular}{@{}lccccccccccccc@{}}
\toprule
Dataset & DTW-1NN & Catch22 & MiniRocket & Hydra & InceptionTime & TS2Vec & TICT & Chronos-2 & TiRex & Moment & NuTime & MantisV2 & TimEE \\
\midrule
PLAID & -- & 0.953 & 0.987 & 0.897 & 0.768 & 0.844 & 0.907 & 0.977 & 0.981 & 0.958 & 0.968 & \textbf{0.989} & 0.965 \\
PhalangesOutlinesCorrect & -- & 0.866 & 0.905 & 0.793 & \textbf{0.912} & 0.859 & 0.852 & 0.904 & 0.905 & 0.905 & 0.844 & 0.889 & 0.904 \\
Phoneme & -- & 0.778 & 0.718 & 0.570 & 0.776 & 0.632 & 0.618 & \textbf{0.841} & 0.831 & 0.789 & 0.789 & 0.828 & 0.823 \\
PickupGestureWiimoteZ & -- & 0.935 & 0.988 & 0.900 & 0.985 & 0.962 & 0.946 & 0.969 & 0.971 & 0.945 & 0.968 & \textbf{0.990} & 0.973 \\
PigAirwayPressure & -- & 0.842 & 0.997 & 0.882 & \textbf{1.000} & 0.977 & 0.687 & 0.845 & 0.922 & 0.578 & 0.925 & 0.979 & 0.734 \\
PigArtPressure & -- & 0.992 & 1.000 & 0.993 & \textbf{1.000} & 0.674 & 0.779 & 0.993 & 0.998 & 0.772 & 0.998 & 0.996 & 0.986 \\
PigCVP & -- & 0.953 & 0.998 & 0.983 & \textbf{0.999} & 0.909 & 0.734 & 0.978 & 0.988 & 0.701 & 0.993 & 0.992 & 0.983 \\
Plane & -- & \textbf{1.000} & \textbf{1.000} & \textbf{1.000} & \textbf{1.000} & 0.999 & \textbf{1.000} & \textbf{1.000} & \textbf{1.000} & 0.999 & \textbf{1.000} & \textbf{1.000} & \textbf{1.000} \\
PowerCons & -- & 0.962 & \textbf{1.000} & 0.978 & 1.000 & 0.992 & 0.979 & 0.988 & 0.995 & 0.967 & 0.992 & 0.991 & 0.999 \\
ProximalPhalanxOutlineAgeGroup & -- & 0.924 & 0.919 & 0.859 & 0.926 & 0.938 & 0.899 & 0.938 & \textbf{0.948} & 0.934 & 0.932 & 0.942 & 0.940 \\
ProximalPhalanxOutlineCorrect & -- & 0.898 & 0.959 & 0.865 & \textbf{0.960} & 0.934 & 0.897 & 0.939 & 0.953 & 0.929 & 0.906 & 0.927 & 0.955 \\
ProximalPhalanxTW & -- & 0.902 & 0.865 & 0.725 & 0.898 & 0.918 & 0.827 & 0.916 & 0.927 & 0.921 & 0.916 & 0.928 & \textbf{0.937} \\
RefrigerationDevices & -- & 0.717 & 0.663 & 0.646 & 0.682 & 0.236 & 0.695 & 0.778 & \textbf{0.789} & 0.705 & 0.722 & 0.726 & 0.699 \\
Rock & -- & 0.941 & 0.946 & 0.957 & 0.891 & 0.763 & 0.831 & 0.984 & 0.985 & 0.942 & 0.886 & 0.952 & \textbf{0.994} \\
ScreenType & -- & 0.707 & 0.655 & 0.620 & \textbf{0.780} & 0.604 & 0.704 & 0.712 & 0.740 & 0.609 & 0.698 & 0.688 & 0.657 \\
SemgHandGenderCh2 & -- & 0.941 & 0.968 & 0.846 & 0.941 & 0.990 & 0.917 & 0.975 & 0.986 & 0.879 & 0.947 & 0.977 & \textbf{0.992} \\
SemgHandMovementCh2 & -- & 0.914 & 0.922 & 0.697 & 0.869 & \textbf{0.985} & 0.866 & 0.939 & 0.943 & 0.858 & 0.927 & 0.949 & 0.974 \\
SemgHandSubjectCh2 & -- & 0.968 & 0.977 & 0.857 & 0.930 & 0.995 & 0.959 & 0.970 & 0.974 & 0.925 & 0.944 & 0.973 & \textbf{0.996} \\
ShakeGestureWiimoteZ & -- & 0.971 & 0.999 & 0.956 & 0.966 & 0.980 & 0.936 & 0.984 & 0.988 & 0.974 & 0.977 & 0.995 & \textbf{1.000} \\
ShapeletSim & -- & 0.999 & \textbf{1.000} & 0.989 & 0.999 & 1.000 & 0.968 & \textbf{1.000} & 1.000 & 0.988 & 0.978 & 0.997 & 0.996 \\
ShapesAll & -- & 0.984 & 0.994 & 0.964 & \textbf{0.997} & 0.992 & 0.959 & 0.995 & 0.991 & 0.990 & 0.988 & 0.993 & 0.996 \\
SmallKitchenAppliances & -- & 0.922 & 0.948 & 0.856 & 0.902 & 0.880 & 0.885 & \textbf{0.949} & 0.941 & 0.865 & 0.934 & 0.948 & 0.933 \\
SmoothSubspace & -- & 0.966 & 0.996 & 0.945 & 0.999 & 0.999 & 0.948 & 0.995 & 0.993 & 0.948 & 0.986 & 0.995 & \textbf{1.000} \\
SonyAIBORobotSurface1 & -- & 0.978 & 0.994 & 0.930 & 0.994 & 0.992 & 0.977 & 0.984 & 0.995 & 0.989 & 0.963 & 0.982 & \textbf{0.996} \\
SonyAIBORobotSurface2 & -- & 0.973 & 0.984 & 0.948 & \textbf{0.991} & 0.968 & 0.958 & 0.957 & 0.959 & 0.964 & 0.918 & 0.987 & 0.976 \\
StarLightCurves & -- & 0.991 & 0.994 & 0.974 & 0.993 & 0.991 & 0.992 & 0.994 & \textbf{0.995} & 0.988 & 0.994 & 0.995 & 0.993 \\
Strawberry & -- & 0.973 & 0.993 & 0.974 & \textbf{0.996} & 0.993 & 0.964 & 0.990 & 0.991 & 0.986 & 0.983 & 0.989 & 0.994 \\
SwedishLeaf & -- & 0.996 & 0.999 & 0.988 & 0.999 & 0.997 & 0.979 & 0.998 & 0.998 & 0.996 & 0.996 & 0.999 & \textbf{0.999} \\
Symbols & -- & 0.997 & 0.996 & 0.988 & \textbf{1.000} & 0.999 & 0.990 & 0.998 & 0.998 & 0.998 & 0.996 & 0.998 & 0.998 \\
SyntheticControl & -- & 1.000 & 1.000 & 0.974 & 1.000 & \textbf{1.000} & 1.000 & 1.000 & 1.000 & 1.000 & 0.999 & 1.000 & 1.000 \\
ToeSegmentation1 & -- & 0.950 & 0.988 & 0.963 & 0.990 & 0.980 & 0.941 & 0.978 & 0.995 & 0.981 & 0.944 & \textbf{0.996} & 0.989 \\
ToeSegmentation2 & -- & 0.856 & 0.979 & 0.879 & \textbf{0.993} & 0.970 & 0.933 & 0.960 & 0.981 & 0.946 & 0.895 & 0.970 & 0.986 \\
Trace & -- & \textbf{1.000} & \textbf{1.000} & \textbf{1.000} & \textbf{1.000} & \textbf{1.000} & \textbf{1.000} & \textbf{1.000} & \textbf{1.000} & \textbf{1.000} & \textbf{1.000} & \textbf{1.000} & \textbf{1.000} \\
TwoLeadECG & -- & 0.929 & \textbf{1.000} & 0.999 & 1.000 & 0.998 & 0.984 & 0.994 & 0.999 & 0.946 & 0.990 & 1.000 & 0.986 \\
TwoPatterns & -- & 0.961 & 1.000 & 0.991 & \textbf{1.000} & \textbf{1.000} & 0.894 & 0.990 & 1.000 & 0.995 & 0.969 & 0.999 & 1.000 \\
UMD & -- & 0.978 & 0.994 & 0.995 & 0.999 & \textbf{1.000} & 0.998 & 0.999 & 0.999 & 0.995 & 0.995 & 0.999 & 0.999 \\
UWaveGestureLibraryAll & -- & 0.977 & 0.998 & 0.985 & 0.995 & 0.995 & 0.991 & 0.994 & 0.995 & 0.991 & 0.987 & 0.989 & \textbf{0.999} \\
UWaveGestureLibraryX & -- & 0.959 & 0.979 & 0.916 & 0.971 & 0.967 & 0.948 & 0.971 & 0.974 & 0.968 & 0.973 & 0.975 & \textbf{0.981} \\
UWaveGestureLibraryY & -- & 0.946 & 0.963 & 0.882 & 0.958 & 0.948 & 0.926 & 0.953 & 0.958 & 0.950 & 0.956 & 0.961 & \textbf{0.968} \\
UWaveGestureLibraryZ & -- & 0.950 & 0.971 & 0.888 & 0.962 & 0.962 & 0.927 & 0.964 & 0.964 & 0.958 & 0.964 & 0.967 & \textbf{0.973} \\
Wafer & -- & 1.000 & 1.000 & \textbf{1.000} & 1.000 & 1.000 & 0.993 & 1.000 & 1.000 & 0.999 & 1.000 & 1.000 & 1.000 \\
Wine & -- & 0.420 & 0.907 & 0.907 & 0.805 & 0.918 & \textbf{0.944} & 0.913 & 0.835 & 0.931 & 0.881 & 0.936 & 0.863 \\
WordSynonyms & -- & 0.892 & 0.939 & 0.786 & 0.956 & 0.921 & 0.804 & 0.932 & 0.916 & 0.933 & 0.885 & 0.935 & \textbf{0.959} \\
Worms & -- & 0.946 & 0.913 & 0.824 & 0.936 & 0.931 & 0.859 & 0.953 & 0.968 & 0.928 & 0.943 & 0.958 & \textbf{0.970} \\
WormsTwoClass & -- & \textbf{0.936} & 0.845 & 0.761 & 0.888 & 0.879 & 0.858 & 0.888 & 0.926 & 0.811 & 0.852 & 0.893 & 0.928 \\
Yoga & -- & 0.866 & 0.968 & 0.926 & \textbf{0.973} & 0.950 & 0.899 & 0.895 & 0.915 & 0.923 & 0.909 & 0.929 & 0.955 \\
\midrule
\textit{Average} & -- & 0.928 & 0.954 & 0.904 & 0.952 & 0.937 & 0.924 & 0.958 & 0.962 & 0.939 & 0.942 & 0.960 & \textbf{0.964} \\
\textit{Mean rank} & nan & 8.69 & 4.77 & 10.08 & 4.77 & 5.94 & 9.67 & 5.62 & 4.32 & 8.23 & 8.00 & 4.41 & \textbf{3.51} \\
\bottomrule
\end{tabular}%
}
\end{table}

\begin{table}[htbp]
\centering
\setlength{\tabcolsep}{4pt}
\caption{UEA multivariate ROCAUC. \textbf{Bold} = best per dataset.}
\label{tab:uea-rocauc}
\resizebox{\linewidth}{!}{%
\begin{tabular}{@{}lcccccccccccccc@{}}
\toprule
Dataset & DTW-1NN & Catch22 & MiniRocket & Hydra & InceptionTime & TS2Vec & TiCT & Chronos-2 & TiRex & Moment & NuTime & MantisV2 & TimEE (VP) & TimEE (PV) \\
\midrule
ArticularyWordRecognition & -- & 0.999 & 1.000 & 0.995 & \textbf{1.000} & 0.999 & 0.999 & 0.999 & 0.999 & 0.999 & 1.000 & 1.000 & 0.999 & 0.998 \\
AtrialFibrillation & -- & 0.323 & 0.227 & 0.350 & 0.413 & 0.303 & 0.260 & 0.453 & 0.560 & 0.427 & 0.410 & 0.360 & 0.353 & \textbf{0.580} \\
BasicMotions & -- & \textbf{1.000} & \textbf{1.000} & \textbf{1.000} & \textbf{1.000} & \textbf{1.000} & \textbf{1.000} & \textbf{1.000} & \textbf{1.000} & 0.997 & \textbf{1.000} & \textbf{1.000} & \textbf{1.000} & \textbf{1.000} \\
CharacterTrajectories & -- & 0.998 & 1.000 & 0.994 & \textbf{1.000} & 1.000 & 0.994 & 0.999 & 0.999 & 1.000 & 1.000 & 0.999 & 1.000 & 0.999 \\
Cricket & -- & 0.998 & 1.000 & 0.992 & 1.000 & 0.999 & 0.999 & 1.000 & \textbf{1.000} & 0.998 & \textbf{1.000} & 1.000 & \textbf{1.000} & 0.999 \\
DuckDuckGeese & -- & 0.794 & \textbf{0.895} & 0.800 & 0.872 & 0.842 & 0.664 & 0.791 & 0.757 & 0.784 & 0.757 & 0.810 & 0.689 & 0.681 \\
ERing & -- & 0.994 & 1.000 & 0.991 & 0.994 & 0.993 & 0.978 & 0.999 & 0.999 & 0.997 & 0.999 & \textbf{1.000} & 0.999 & 0.997 \\
Epilepsy & -- & 0.999 & \textbf{1.000} & \textbf{1.000} & 0.999 & 0.998 & 0.998 & \textbf{1.000} & \textbf{1.000} & 1.000 & \textbf{1.000} & \textbf{1.000} & \textbf{1.000} & \textbf{1.000} \\
EthanolConcentration & -- & 0.536 & 0.697 & 0.699 & 0.522 & 0.628 & 0.520 & 0.609 & 0.559 & 0.509 & 0.680 & 0.674 & 0.831 & \textbf{0.833} \\
FingerMovements & -- & 0.539 & 0.602 & 0.552 & 0.567 & 0.532 & 0.549 & 0.550 & 0.559 & 0.587 & 0.546 & 0.565 & \textbf{0.632} & 0.606 \\
HandMovementDirection & -- & 0.499 & 0.642 & 0.483 & \textbf{0.679} & 0.563 & 0.610 & 0.526 & 0.551 & 0.550 & 0.602 & 0.552 & 0.568 & 0.629 \\
Handwriting & -- & 0.799 & 0.899 & 0.721 & \textbf{0.951} & 0.843 & 0.729 & 0.780 & 0.770 & 0.800 & 0.738 & 0.810 & 0.827 & 0.837 \\
Heartbeat & -- & 0.641 & \textbf{0.799} & 0.624 & 0.767 & 0.729 & 0.628 & 0.750 & 0.750 & 0.740 & 0.784 & 0.790 & 0.761 & 0.738 \\
JapaneseVowels & -- & 0.961 & 1.000 & 0.987 & \textbf{1.000} & 0.997 & 0.915 & 0.970 & 0.968 & 0.959 & 0.999 & 0.998 & 0.981 & 0.968 \\
LSST & -- & 0.878 & 0.873 & 0.684 & 0.743 & 0.905 & 0.801 & 0.880 & 0.854 & 0.837 & 0.874 & \textbf{0.906} & 0.861 & 0.881 \\
Libras & -- & 0.991 & 0.995 & 0.964 & 0.990 & 0.991 & 0.961 & 0.993 & 0.994 & 0.985 & 0.995 & \textbf{0.996} & 0.994 & 0.991 \\
MotorImagery & -- & 0.463 & 0.561 & 0.500 & 0.595 & 0.542 & 0.541 & \textbf{0.633} & 0.559 & 0.595 & 0.601 & 0.505 & 0.478 & 0.582 \\
NATOPS & -- & 0.973 & 0.991 & 0.923 & \textbf{0.997} & 0.990 & 0.941 & 0.970 & 0.970 & 0.972 & 0.976 & 0.983 & 0.977 & 0.968 \\
PEMS-SF & -- & 1.000 & 0.967 & 0.882 & 0.970 & 1.000 & 0.993 & \textbf{1.000} & \textbf{1.000} & \textbf{1.000} & \textbf{1.000} & \textbf{1.000} & 0.995 & 0.997 \\
RacketSports & -- & 0.952 & 0.966 & 0.946 & 0.975 & 0.962 & 0.931 & 0.956 & 0.956 & 0.945 & \textbf{0.978} & 0.977 & 0.973 & 0.967 \\
SelfRegulationSCP1 & -- & 0.790 & \textbf{0.971} & 0.867 & 0.944 & 0.907 & 0.803 & 0.895 & 0.900 & 0.854 & 0.926 & 0.948 & 0.957 & 0.959 \\
SelfRegulationSCP2 & -- & 0.581 & 0.541 & \textbf{0.589} & 0.485 & 0.526 & 0.461 & 0.530 & 0.509 & 0.459 & 0.520 & 0.534 & 0.426 & 0.472 \\
StandWalkJump & -- & 0.530 & 0.693 & 0.550 & 0.653 & 0.377 & \textbf{0.880} & 0.807 & 0.767 & 0.843 & 0.653 & 0.587 & 0.580 & 0.580 \\
UWaveGestureLibrary & -- & 0.980 & 0.992 & 0.948 & 0.988 & 0.991 & 0.950 & 0.988 & 0.986 & \textbf{0.992} & 0.986 & 0.990 & 0.990 & 0.986 \\
\midrule
\textit{Average} & -- & 0.801 & \textbf{0.846} & 0.793 & 0.838 & 0.817 & 0.796 & 0.837 & 0.832 & 0.826 & 0.834 & 0.833 & 0.828 & 0.844 \\
\textit{Mean rank} & nan & 9.31 & \textbf{4.12} & 10.00 & 5.48 & 7.29 & 10.65 & 6.33 & 6.77 & 7.92 & 5.31 & 4.75 & 6.29 & 6.77 \\
\bottomrule
\end{tabular}%
}
\end{table}

\begin{table}[htbp]
\centering
\setlength{\tabcolsep}{4pt}
\caption{UCR univariate accuracy. \textbf{Bold} = best per dataset. Best variant shown for models with multiple checkpoints.}
\label{tab:ucr-accuracy}
\resizebox{\linewidth}{!}{%
\begin{tabular}{@{}lccccccccccccc@{}}
\toprule
Dataset & DTW-1NN & Catch22 & MiniRocket & Hydra & InceptionTime & TS2Vec & TICT & Chronos-2 & TiRex & Moment & NuTime & MantisV2 & TimEE \\
\midrule
ACSF1 & 0.620 & 0.830 & 0.910 & 0.850 & \textbf{0.930} & 0.890 & 0.400 & 0.850 & 0.890 & 0.820 & 0.760 & 0.830 & 0.830 \\
Adiac & 0.609 & 0.719 & 0.813 & 0.818 & 0.831 & 0.754 & 0.481 & 0.831 & 0.788 & 0.772 & 0.749 & \textbf{0.844} & 0.829 \\
AllGestureWiimoteX & 0.717 & 0.529 & 0.729 & 0.703 & \textbf{0.774} & 0.726 & 0.570 & 0.586 & 0.604 & 0.590 & 0.586 & 0.633 & 0.669 \\
AllGestureWiimoteY & 0.730 & 0.613 & 0.753 & 0.731 & \textbf{0.789} & 0.721 & 0.583 & 0.639 & 0.684 & 0.647 & 0.543 & 0.677 & 0.729 \\
AllGestureWiimoteZ & 0.651 & 0.521 & 0.736 & 0.686 & \textbf{0.806} & 0.709 & 0.493 & 0.601 & 0.590 & 0.587 & 0.541 & 0.691 & 0.697 \\
ArrowHead & 0.800 & 0.760 & \textbf{0.869} & 0.834 & 0.846 & 0.829 & 0.766 & 0.829 & 0.829 & 0.823 & 0.777 & 0.817 & 0.823 \\
BME & 0.980 & 0.933 & \textbf{1.000} & \textbf{1.000} & \textbf{1.000} & 0.987 & 0.687 & 0.987 & \textbf{1.000} & 0.967 & 0.840 & \textbf{1.000} & \textbf{1.000} \\
Beef & 0.667 & 0.600 & 0.833 & 0.833 & 0.667 & 0.667 & 0.600 & 0.733 & \textbf{0.900} & 0.733 & 0.700 & 0.700 & \textbf{0.900} \\
BeetleFly & 0.700 & 0.750 & 0.900 & 0.900 & 0.850 & 0.850 & 0.850 & 0.850 & 0.950 & \textbf{1.000} & 0.850 & 0.950 & 0.850 \\
BirdChicken & 0.700 & 0.850 & 0.900 & 0.900 & \textbf{0.950} & 0.800 & 0.900 & 0.900 & 0.900 & 0.900 & \textbf{0.950} & 0.900 & 0.900 \\
CBF & 0.996 & 0.961 & \textbf{0.999} & 0.993 & \textbf{0.999} & \textbf{0.999} & 0.942 & \textbf{0.999} & 0.998 & 0.967 & 0.976 & 0.997 & \textbf{0.999} \\
Car & 0.767 & 0.783 & 0.917 & \textbf{0.933} & 0.883 & 0.683 & 0.617 & 0.767 & 0.833 & 0.783 & 0.767 & 0.767 & 0.900 \\
Chinatown & 0.953 & 0.921 & 0.983 & 0.983 & \textbf{0.985} & 0.959 & 0.930 & 0.980 & 0.974 & 0.980 & 0.936 & 0.950 & \textbf{0.985} \\
ChlorineConcentration & 0.650 & 0.599 & 0.758 & 0.760 & \textbf{0.855} & 0.825 & 0.601 & 0.714 & 0.749 & 0.709 & 0.676 & 0.704 & 0.609 \\
CinCECGTorso & 0.930 & 0.808 & 0.868 & \textbf{0.995} & 0.845 & 0.775 & 0.643 & 0.926 & 0.992 & 0.605 & 0.740 & 0.812 & 0.989 \\
Coffee & \textbf{1.000} & \textbf{1.000} & \textbf{1.000} & \textbf{1.000} & \textbf{1.000} & \textbf{1.000} & 0.964 & \textbf{1.000} & \textbf{1.000} & \textbf{1.000} & 0.964 & \textbf{1.000} & \textbf{1.000} \\
Computers & 0.620 & 0.724 & 0.732 & 0.716 & \textbf{0.832} & 0.660 & 0.688 & 0.744 & 0.752 & 0.668 & 0.772 & 0.736 & 0.656 \\
CricketX & 0.772 & 0.579 & 0.808 & 0.800 & \textbf{0.841} & 0.797 & 0.603 & 0.708 & 0.697 & 0.687 & 0.667 & 0.777 & 0.810 \\
CricketY & 0.759 & 0.546 & 0.828 & 0.841 & \textbf{0.862} & 0.756 & 0.579 & 0.721 & 0.754 & 0.667 & 0.692 & 0.790 & 0.851 \\
CricketZ & 0.746 & 0.636 & 0.826 & 0.805 & \textbf{0.862} & 0.815 & 0.610 & 0.751 & 0.726 & 0.713 & 0.685 & 0.818 & 0.841 \\
Crop & 0.712 & 0.651 & 0.763 & 0.734 & \textbf{0.797} & 0.750 & 0.656 & 0.736 & 0.747 & 0.711 & 0.672 & 0.736 & 0.784 \\
DiatomSizeReduction & 0.935 & 0.941 & 0.928 & 0.948 & 0.941 & 0.967 & 0.869 & 0.899 & 0.869 & 0.866 & 0.850 & 0.843 & \textbf{0.987} \\
DistalPhalanxOutlineAgeGroup & 0.626 & 0.712 & 0.748 & 0.763 & 0.719 & 0.727 & 0.719 & 0.763 & 0.755 & 0.734 & 0.734 & 0.763 & \textbf{0.791} \\
DistalPhalanxOutlineCorrect & 0.725 & 0.783 & 0.761 & 0.786 & 0.757 & 0.732 & 0.732 & \textbf{0.812} & 0.793 & 0.786 & 0.775 & 0.793 & 0.786 \\
DistalPhalanxTW & 0.633 & 0.669 & 0.669 & \textbf{0.712} & 0.683 & 0.676 & 0.683 & 0.698 & 0.683 & 0.691 & 0.691 & \textbf{0.712} & 0.676 \\
DodgerLoopDay & 0.588 & 0.519 & \textbf{0.688} & 0.519 & 0.571 & 0.662 & 0.532 & 0.558 & 0.623 & 0.442 & 0.545 & 0.558 & 0.649 \\
DodgerLoopGame & \textbf{0.927} & 0.693 & 0.850 & 0.819 & 0.858 & 0.921 & 0.858 & 0.866 & 0.843 & 0.764 & 0.787 & 0.795 & 0.913 \\
DodgerLoopWeekend & 0.978 & 0.944 & \textbf{0.984} & 0.976 & \textbf{0.984} & \textbf{0.984} & 0.944 & 0.952 & 0.952 & 0.937 & 0.968 & 0.952 & \textbf{0.984} \\
ECG200 & 0.880 & 0.820 & 0.920 & 0.870 & \textbf{0.930} & 0.910 & 0.790 & 0.860 & 0.860 & 0.880 & 0.830 & 0.830 & 0.920 \\
ECG5000 & 0.925 & 0.938 & 0.945 & \textbf{0.949} & 0.941 & 0.935 & 0.927 & 0.935 & 0.940 & 0.943 & 0.934 & 0.939 & 0.949 \\
ECGFiveDays & 0.797 & 0.777 & \textbf{1.000} & \textbf{1.000} & \textbf{1.000} & \textbf{1.000} & 0.861 & 0.941 & 0.980 & 0.846 & 0.815 & 0.930 & 0.994 \\
EOGHorizontalSignal & 0.475 & 0.552 & 0.605 & 0.572 & 0.616 & 0.506 & 0.403 & 0.533 & 0.561 & 0.492 & 0.439 & 0.608 & \textbf{0.635} \\
EOGVerticalSignal & 0.475 & 0.456 & 0.519 & 0.492 & 0.486 & 0.506 & 0.390 & 0.401 & 0.470 & 0.392 & 0.301 & 0.483 & \textbf{0.547} \\
Earthquakes & 0.727 & 0.748 & 0.748 & 0.734 & 0.748 & 0.748 & \textbf{0.755} & 0.741 & 0.748 & \textbf{0.755} & 0.741 & 0.748 & 0.748 \\
ElectricDevices & 0.619 & 0.725 & 0.738 & \textbf{0.752} & 0.708 & 0.731 & 0.629 & 0.749 & 0.713 & 0.647 & 0.707 & 0.745 & 0.609 \\
EthanolLevel & 0.282 & 0.352 & 0.596 & 0.588 & \textbf{0.822} & 0.454 & 0.340 & 0.420 & 0.672 & 0.400 & 0.352 & 0.360 & 0.644 \\
FaceAll & 0.808 & 0.764 & 0.807 & \textbf{0.898} & 0.796 & 0.777 & 0.739 & 0.742 & 0.881 & 0.684 & 0.652 & 0.719 & 0.802 \\
FaceFour & 0.886 & 0.648 & 0.989 & 0.886 & 0.955 & 0.886 & 0.784 & 0.716 & 0.852 & 0.795 & 0.830 & 0.909 & \textbf{1.000} \\
FacesUCR & 0.912 & 0.695 & 0.958 & 0.956 & \textbf{0.962} & 0.929 & 0.684 & 0.783 & 0.849 & 0.727 & 0.713 & 0.844 & 0.941 \\
FiftyWords & 0.758 & 0.602 & 0.840 & \textbf{0.842} & 0.791 & 0.765 & 0.587 & 0.670 & 0.701 & 0.714 & 0.613 & 0.719 & 0.811 \\
Fish & 0.846 & 0.749 & 0.977 & \textbf{0.989} & 0.983 & 0.920 & 0.754 & 0.874 & 0.920 & 0.869 & 0.926 & 0.937 & 0.926 \\
FordA & 0.691 & 0.920 & 0.951 & \textbf{0.961} & \textbf{0.961} & 0.927 & 0.898 & 0.918 & 0.958 & 0.898 & 0.900 & 0.930 & 0.949 \\
FordB & 0.607 & 0.752 & 0.804 & 0.825 & \textbf{0.859} & 0.798 & 0.723 & 0.791 & 0.838 & 0.772 & 0.756 & 0.800 & 0.802 \\
FreezerRegularTrain & 0.907 & 0.998 & \textbf{1.000} & 0.998 & 0.997 & 0.984 & 0.896 & 0.972 & 0.994 & 0.898 & 0.978 & 0.984 & 0.998 \\
FreezerSmallTrain & 0.676 & 0.950 & \textbf{0.968} & 0.926 & 0.840 & 0.885 & 0.807 & 0.925 & 0.919 & 0.755 & 0.945 & 0.930 & 0.943 \\
Fungi & 0.823 & 0.903 & \textbf{1.000} & \textbf{1.000} & 0.995 & 0.962 & 0.796 & 0.935 & 0.930 & 0.968 & 0.742 & 0.930 & 0.941 \\
GestureMidAirD1 & 0.639 & 0.746 & 0.738 & 0.754 & 0.669 & 0.485 & 0.646 & 0.754 & 0.746 & 0.769 & 0.562 & 0.700 & \textbf{0.785} \\
GestureMidAirD2 & 0.600 & 0.662 & 0.692 & 0.685 & 0.508 & 0.392 & 0.577 & 0.654 & \textbf{0.708} & 0.700 & 0.485 & \textbf{0.708} & 0.685 \\
GestureMidAirD3 & 0.377 & 0.469 & 0.446 & 0.485 & 0.246 & 0.200 & 0.338 & \textbf{0.554} & 0.500 & 0.462 & 0.315 & 0.400 & \textbf{0.554} \\
GesturePebbleZ1 & 0.826 & 0.779 & \textbf{0.907} & 0.890 & 0.855 & 0.692 & 0.767 & 0.890 & 0.890 & 0.872 & 0.791 & 0.884 & 0.860 \\
GesturePebbleZ2 & 0.778 & 0.753 & 0.899 & 0.886 & 0.835 & 0.722 & 0.759 & 0.873 & \textbf{0.905} & 0.880 & 0.785 & 0.880 & 0.816 \\
GunPoint & 0.913 & 0.953 & 0.993 & \textbf{1.000} & \textbf{1.000} & 0.980 & 0.940 & 0.993 & 0.980 & 0.973 & 0.960 & 0.980 & 0.987 \\
GunPointAgeSpan & 0.965 & 0.956 & 0.997 & \textbf{1.000} & 0.978 & 0.965 & 0.972 & 0.987 & 0.975 & 0.946 & 0.972 & 0.994 & 0.991 \\
GunPointMaleVersusFemale & 0.975 & 0.994 & \textbf{1.000} & \textbf{1.000} & 0.997 & 0.997 & 0.997 & 0.991 & 0.994 & 0.984 & 0.972 & \textbf{1.000} & \textbf{1.000} \\
GunPointOldVersusYoung & 0.965 & 0.975 & \textbf{1.000} & \textbf{1.000} & \textbf{1.000} & \textbf{1.000} & 0.987 & 0.997 & \textbf{1.000} & 0.975 & \textbf{1.000} & 0.997 & \textbf{1.000} \\
Ham & 0.600 & 0.600 & 0.686 & 0.743 & 0.705 & 0.676 & 0.667 & 0.705 & \textbf{0.762} & 0.695 & 0.733 & 0.676 & 0.686 \\
HandOutlines & 0.862 & 0.854 & 0.943 & 0.941 & \textbf{0.957} & 0.916 & 0.886 & 0.914 & 0.900 & 0.905 & 0.892 & 0.903 & 0.932 \\
Haptics & 0.412 & 0.445 & 0.526 & 0.513 & \textbf{0.558} & 0.523 & 0.451 & 0.526 & 0.526 & 0.464 & 0.461 & 0.497 & 0.552 \\
Herring & 0.531 & 0.578 & 0.688 & \textbf{0.750} & 0.703 & 0.594 & 0.656 & 0.688 & 0.625 & 0.609 & 0.625 & 0.672 & 0.578 \\
HouseTwenty & 0.941 & 0.958 & \textbf{0.966} & 0.958 & 0.924 & 0.908 & 0.899 & 0.950 & \textbf{0.966} & 0.916 & 0.866 & 0.941 & 0.958 \\
InlineSkate & 0.387 & 0.451 & 0.476 & \textbf{0.491} & 0.473 & 0.424 & 0.384 & 0.433 & 0.442 & 0.338 & 0.336 & 0.467 & 0.442 \\
InsectEPGRegularTrain & 0.827 & 0.960 & \textbf{1.000} & \textbf{1.000} & \textbf{1.000} & \textbf{1.000} & 0.900 & 0.996 & \textbf{1.000} & 0.928 & \textbf{1.000} & \textbf{1.000} & 0.964 \\
InsectEPGSmallTrain & 0.695 & 0.827 & \textbf{1.000} & \textbf{1.000} & \textbf{1.000} & \textbf{1.000} & 0.795 & 0.884 & 0.968 & 0.871 & \textbf{1.000} & \textbf{1.000} & 0.956 \\
InsectWingbeatSound & 0.585 & 0.566 & 0.669 & 0.655 & 0.631 & 0.617 & 0.494 & 0.653 & \textbf{0.675} & 0.627 & 0.526 & 0.603 & 0.655 \\
ItalyPowerDemand & 0.955 & 0.881 & 0.966 & 0.966 & \textbf{0.969} & 0.964 & 0.867 & 0.958 & 0.964 & 0.956 & 0.882 & 0.927 & 0.961 \\
LargeKitchenAppliances & 0.795 & 0.827 & 0.856 & 0.883 & \textbf{0.896} & 0.861 & 0.717 & 0.888 & 0.813 & 0.760 & 0.760 & 0.763 & 0.675 \\
Lightning2 & \textbf{0.869} & 0.705 & 0.754 & 0.721 & 0.787 & 0.869 & 0.754 & 0.721 & 0.738 & 0.754 & 0.738 & 0.721 & 0.852 \\
Lightning7 & 0.712 & 0.671 & 0.795 & 0.808 & \textbf{0.849} & 0.822 & 0.712 & 0.699 & 0.726 & 0.712 & 0.699 & 0.685 & \textbf{0.849} \\
Mallat & 0.914 & 0.899 & 0.946 & 0.941 & 0.949 & 0.887 & 0.501 & 0.909 & \textbf{0.959} & 0.897 & 0.850 & 0.900 & 0.924 \\
Meat & 0.933 & 0.867 & 0.967 & 0.917 & 0.933 & 0.933 & 0.617 & 0.933 & 0.917 & 0.950 & 0.917 & 0.933 & \textbf{0.983} \\
MedicalImages & 0.747 & 0.762 & 0.797 & 0.778 & 0.795 & \textbf{0.812} & 0.758 & 0.729 & 0.763 & 0.713 & 0.714 & 0.759 & 0.796 \\
MelbournePedestrian & 0.816 & 0.804 & 0.969 & 0.931 & \textbf{0.975} & 0.964 & 0.805 & 0.945 & 0.953 & 0.865 & 0.925 & 0.953 & 0.931 \\
MiddlePhalanxOutlineAgeGroup & 0.520 & 0.617 & 0.571 & 0.597 & 0.513 & 0.649 & 0.578 & 0.604 & 0.604 & 0.584 & \textbf{0.656} & 0.591 & 0.636 \\
MiddlePhalanxOutlineCorrect & 0.766 & 0.766 & 0.838 & 0.835 & 0.842 & 0.825 & 0.784 & \textbf{0.863} & 0.859 & 0.845 & 0.787 & 0.835 & 0.852 \\
MiddlePhalanxTW & 0.506 & 0.532 & 0.526 & 0.539 & 0.552 & 0.584 & 0.545 & 0.552 & 0.539 & 0.584 & 0.552 & 0.532 & \textbf{0.591} \\
MixedShapesRegularTrain & 0.909 & 0.918 & 0.969 & \textbf{0.981} & 0.967 & 0.920 & 0.919 & 0.942 & 0.964 & 0.890 & 0.938 & 0.948 & 0.979 \\
MixedShapesSmallTrain & 0.833 & 0.864 & 0.951 & \textbf{0.965} & 0.916 & 0.855 & 0.797 & 0.909 & 0.943 & 0.819 & 0.917 & 0.921 & 0.955 \\
MoteStrain & 0.866 & 0.867 & 0.932 & 0.929 & 0.897 & 0.860 & 0.899 & 0.926 & 0.922 & 0.903 & \textbf{0.942} & 0.934 & 0.900 \\
NonInvasiveFetalECGThorax1 & 0.811 & 0.852 & 0.945 & 0.943 & \textbf{0.963} & 0.937 & 0.769 & 0.867 & 0.918 & 0.873 & 0.782 & 0.862 & 0.931 \\
NonInvasiveFetalECGThorax2 & 0.871 & 0.882 & \textbf{0.964} & 0.958 & 0.962 & 0.940 & 0.813 & 0.892 & 0.931 & 0.909 & 0.820 & 0.887 & 0.942 \\
OSULeaf & 0.612 & 0.707 & 0.959 & \textbf{0.988} & 0.950 & 0.843 & 0.727 & 0.901 & 0.938 & 0.748 & 0.798 & 0.934 & 0.868 \\
OliveOil & 0.867 & 0.700 & 0.933 & 0.933 & 0.800 & 0.900 & 0.400 & 0.933 & 0.933 & 0.900 & 0.733 & 0.833 & \textbf{0.967} \\
PLAID & 0.834 & 0.767 & \textbf{0.922} & 0.873 & 0.467 & 0.540 & 0.667 & 0.844 & 0.838 & 0.780 & 0.793 & 0.886 & 0.844 \\
PhalangesOutlinesCorrect & 0.761 & 0.784 & 0.837 & 0.823 & 0.838 & 0.772 & 0.784 & 0.836 & \textbf{0.841} & 0.838 & 0.784 & 0.829 & 0.824 \\
Phoneme & 0.227 & 0.309 & 0.285 & 0.324 & 0.352 & 0.310 & 0.183 & 0.399 & 0.389 & 0.312 & 0.294 & 0.371 & \textbf{0.410} \\
PickupGestureWiimoteZ & 0.660 & 0.540 & 0.880 & 0.820 & \textbf{0.920} & 0.740 & 0.700 & 0.800 & 0.820 & 0.660 & 0.740 & 0.840 & 0.780 \\
PigAirwayPressure & 0.096 & 0.308 & 0.870 & 0.769 & \textbf{0.947} & 0.644 & 0.139 & 0.351 & 0.399 & 0.106 & 0.389 & 0.625 & 0.130 \\
PigArtPressure & 0.197 & 0.909 & 0.986 & 0.986 & \textbf{1.000} & 0.971 & 0.332 & 0.812 & 0.923 & 0.173 & 0.942 & 0.933 & 0.639 \\
PigCVP & 0.159 & 0.538 & 0.947 & \textbf{0.966} & 0.952 & 0.861 & 0.250 & 0.745 & 0.851 & 0.120 & 0.832 & 0.861 & 0.731 \\
\bottomrule
\end{tabular}%
}
\end{table}

\begin{table}[htbp]
\centering
\setlength{\tabcolsep}{4pt}
\caption{UCR univariate accuracy. \textbf{Bold} = best per dataset. Best variant shown for models with multiple checkpoints. (continued)}
\label{tab:ucr-accuracy-cont}
\resizebox{\linewidth}{!}{%
\begin{tabular}{@{}lccccccccccccc@{}}
\toprule
Dataset & DTW-1NN & Catch22 & MiniRocket & Hydra & InceptionTime & TS2Vec & TICT & Chronos-2 & TiRex & Moment & NuTime & MantisV2 & TimEE \\
\midrule
Plane & \textbf{1.000} & \textbf{1.000} & \textbf{1.000} & \textbf{1.000} & \textbf{1.000} & 0.971 & \textbf{1.000} & \textbf{1.000} & \textbf{1.000} & 0.981 & \textbf{1.000} & \textbf{1.000} & \textbf{1.000} \\
PowerCons & 0.922 & 0.917 & 0.989 & 0.978 & \textbf{0.994} & 0.961 & 0.917 & 0.922 & 0.978 & 0.856 & 0.933 & 0.972 & 0.989 \\
ProximalPhalanxOutlineAgeGroup & 0.785 & 0.834 & 0.844 & \textbf{0.868} & 0.849 & 0.834 & 0.771 & 0.859 & 0.863 & 0.849 & 0.849 & 0.834 & 0.863 \\
ProximalPhalanxOutlineCorrect & 0.790 & 0.838 & 0.904 & 0.911 & \textbf{0.914} & 0.890 & 0.832 & 0.859 & 0.904 & 0.883 & 0.838 & 0.869 & 0.907 \\
ProximalPhalanxTW & 0.756 & 0.766 & 0.815 & 0.810 & 0.771 & 0.795 & 0.776 & 0.805 & 0.820 & 0.805 & 0.795 & \textbf{0.824} & 0.805 \\
RefrigerationDevices & 0.440 & 0.517 & 0.496 & 0.528 & 0.531 & \textbf{0.616} & 0.528 & 0.560 & 0.581 & 0.528 & 0.533 & 0.557 & 0.501 \\
Rock & 0.840 & 0.740 & 0.820 & 0.920 & 0.620 & 0.760 & 0.640 & \textbf{0.940} & \textbf{0.940} & 0.760 & 0.720 & 0.820 & \textbf{0.940} \\
ScreenType & 0.411 & 0.533 & 0.475 & 0.493 & \textbf{0.589} & 0.413 & 0.544 & 0.507 & 0.541 & 0.429 & 0.507 & 0.485 & 0.448 \\
SemgHandGenderCh2 & 0.845 & 0.877 & 0.900 & 0.847 & 0.875 & \textbf{0.958} & 0.825 & 0.920 & 0.950 & 0.810 & 0.865 & 0.920 & 0.957 \\
SemgHandMovementCh2 & 0.638 & 0.662 & 0.691 & 0.496 & 0.558 & \textbf{0.876} & 0.587 & 0.718 & 0.709 & 0.553 & 0.680 & 0.753 & 0.813 \\
SemgHandSubjectCh2 & 0.800 & 0.807 & 0.860 & 0.771 & 0.682 & \textbf{0.951} & 0.820 & 0.844 & 0.867 & 0.724 & 0.733 & 0.831 & 0.940 \\
ShakeGestureWiimoteZ & 0.840 & 0.800 & 0.920 & 0.920 & 0.840 & 0.900 & 0.660 & 0.920 & 0.860 & 0.820 & 0.860 & 0.940 & \textbf{0.960} \\
ShapeletSim & 0.700 & 0.972 & \textbf{1.000} & 0.989 & 0.983 & 0.994 & 0.900 & \textbf{1.000} & 0.967 & 0.928 & 0.928 & 0.972 & 0.967 \\
ShapesAll & 0.802 & 0.777 & 0.922 & \textbf{0.928} & 0.925 & 0.903 & 0.698 & 0.883 & 0.857 & 0.830 & 0.842 & 0.875 & 0.900 \\
SmallKitchenAppliances & 0.672 & 0.805 & \textbf{0.837} & 0.808 & 0.792 & 0.699 & 0.720 & 0.827 & 0.832 & 0.691 & 0.813 & 0.835 & 0.803 \\
SmoothSubspace & 0.947 & 0.853 & 0.953 & 0.927 & 0.973 & 0.980 & 0.833 & 0.973 & 0.953 & 0.793 & 0.907 & 0.947 & \textbf{0.993} \\
SonyAIBORobotSurface1 & 0.696 & 0.845 & 0.894 & 0.920 & 0.869 & 0.913 & 0.894 & 0.709 & 0.902 & 0.779 & 0.772 & 0.802 & \textbf{0.923} \\
SonyAIBORobotSurface2 & 0.859 & 0.917 & 0.914 & 0.948 & \textbf{0.951} & 0.908 & 0.887 & 0.879 & 0.852 & 0.888 & 0.838 & 0.937 & 0.916 \\
StarLightCurves & 0.905 & 0.970 & \textbf{0.982} & 0.982 & 0.980 & 0.971 & 0.978 & 0.980 & 0.979 & 0.955 & 0.979 & 0.981 & 0.982 \\
Strawberry & 0.946 & 0.927 & 0.981 & 0.973 & \textbf{0.984} & 0.968 & 0.911 & 0.959 & 0.957 & 0.946 & 0.935 & 0.965 & 0.976 \\
SwedishLeaf & 0.846 & 0.890 & 0.965 & 0.978 & 0.965 & 0.931 & 0.819 & 0.944 & 0.946 & 0.914 & 0.930 & 0.955 & \textbf{0.979} \\
Symbols & 0.938 & 0.958 & \textbf{0.983} & 0.980 & \textbf{0.983} & 0.979 & 0.920 & 0.979 & 0.970 & 0.969 & 0.942 & 0.969 & 0.966 \\
SyntheticControl & 0.983 & 0.987 & 0.987 & 0.957 & 0.997 & \textbf{1.000} & 0.977 & 0.997 & 0.987 & 0.990 & 0.973 & 0.990 & 0.983 \\
ToeSegmentation1 & 0.750 & 0.860 & 0.961 & 0.961 & \textbf{0.974} & 0.917 & 0.860 & 0.921 & 0.947 & 0.934 & 0.860 & 0.965 & 0.939 \\
ToeSegmentation2 & 0.908 & 0.777 & 0.908 & 0.908 & \textbf{0.946} & 0.900 & 0.862 & 0.854 & 0.923 & 0.892 & 0.746 & 0.869 & 0.892 \\
Trace & 0.990 & \textbf{1.000} & \textbf{1.000} & \textbf{1.000} & \textbf{1.000} & \textbf{1.000} & \textbf{1.000} & \textbf{1.000} & \textbf{1.000} & \textbf{1.000} & 0.990 & \textbf{1.000} & \textbf{1.000} \\
TwoLeadECG & 0.868 & 0.818 & 0.998 & \textbf{0.999} & 0.996 & 0.977 & 0.948 & 0.955 & 0.992 & 0.847 & 0.930 & 0.995 & 0.946 \\
TwoPatterns & 0.999 & 0.846 & 0.996 & 0.987 & \textbf{1.000} & \textbf{1.000} & 0.706 & 0.922 & 0.993 & 0.947 & 0.846 & 0.982 & 0.996 \\
UMD & 0.972 & 0.896 & \textbf{0.993} & \textbf{0.993} & \textbf{0.993} & 0.986 & 0.972 & 0.986 & 0.965 & 0.965 & 0.951 & 0.986 & 0.965 \\
UWaveGestureLibraryAll & 0.966 & 0.831 & 0.973 & 0.973 & 0.934 & 0.936 & 0.912 & 0.944 & 0.942 & 0.910 & 0.889 & 0.886 & \textbf{0.979} \\
UWaveGestureLibraryX & 0.773 & 0.761 & 0.848 & \textbf{0.854} & 0.822 & 0.802 & 0.774 & 0.807 & 0.808 & 0.803 & 0.808 & 0.809 & 0.849 \\
UWaveGestureLibraryY & 0.699 & 0.707 & 0.776 & 0.792 & 0.762 & 0.720 & 0.720 & 0.738 & 0.748 & 0.725 & 0.742 & 0.759 & \textbf{0.793} \\
UWaveGestureLibraryZ & 0.678 & 0.707 & \textbf{0.803} & 0.803 & 0.775 & 0.762 & 0.712 & 0.745 & 0.741 & 0.742 & 0.756 & 0.758 & 0.784 \\
Wafer & 0.996 & 0.997 & 0.999 & \textbf{1.000} & 0.999 & 0.998 & 0.984 & 0.991 & 0.998 & 0.990 & 0.994 & 0.996 & 0.998 \\
Wine & 0.611 & 0.463 & 0.852 & 0.907 & 0.685 & \textbf{0.926} & 0.889 & 0.889 & 0.759 & 0.833 & 0.796 & 0.889 & 0.778 \\
WordSynonyms & 0.738 & 0.538 & 0.755 & 0.743 & 0.716 & 0.693 & 0.513 & 0.569 & 0.574 & 0.596 & 0.539 & 0.616 & \textbf{0.757} \\
Worms & 0.532 & 0.753 & 0.753 & 0.753 & 0.766 & 0.753 & 0.688 & 0.779 & \textbf{0.805} & 0.675 & 0.740 & 0.792 & \textbf{0.805} \\
WormsTwoClass & 0.584 & 0.857 & 0.766 & 0.766 & 0.766 & 0.792 & 0.818 & 0.805 & 0.844 & 0.727 & 0.792 & 0.857 & \textbf{0.870} \\
Yoga & 0.844 & 0.771 & 0.913 & \textbf{0.927} & 0.912 & 0.895 & 0.809 & 0.809 & 0.820 & 0.840 & 0.827 & 0.838 & 0.879 \\
\midrule
\textit{Average} & 0.802 & 0.815 & 0.883 & 0.879 & 0.863 & 0.878 & 0.810 & 0.864 & 0.874 & 0.823 & 0.830 & 0.873 & \textbf{0.891} \\
\textit{Mean rank} & 10.21 & 9.73 & \textbf{4.37} & 4.54 & 5.10 & 5.94 & 10.01 & 6.38 & 5.71 & 9.55 & 9.27 & 5.69 & 4.50 \\
\bottomrule
\end{tabular}%
}
\end{table}

\begin{table}[htbp]
\centering
\setlength{\tabcolsep}{4pt}
\caption{UEA multivariate accuracy. \textbf{Bold} = best per dataset. Best variant shown for models with multiple checkpoints.}
\label{tab:uea-accuracy}
\resizebox{\linewidth}{!}{%
\begin{tabular}{@{}lcccccccccccccc@{}}
\toprule
Dataset & DTW-1NN & Catch22 & MiniRocket & Hydra & InceptionTime & TS2Vec & TiCT & Chronos-2 & TiRex & Moment & NuTime & MantisV2 & TimEE (VP) & TimEE (PV) \\
\midrule
ArticularyWordRecognition & 0.987 & 0.983 & \textbf{0.993} & 0.990 & 0.987 & 0.990 & 0.970 & \textbf{0.993} & 0.990 & 0.990 & \textbf{0.993} & \textbf{0.993} & 0.990 & 0.980 \\
AtrialFibrillation & 0.200 & 0.200 & 0.133 & 0.133 & 0.200 & 0.200 & 0.267 & \textbf{0.333} & 0.267 & 0.267 & 0.267 & 0.200 & 0.133 & \textbf{0.333} \\
BasicMotions & 0.975 & 0.950 & \textbf{1.000} & \textbf{1.000} & \textbf{1.000} & \textbf{1.000} & \textbf{1.000} & \textbf{1.000} & \textbf{1.000} & 0.950 & \textbf{1.000} & \textbf{1.000} & \textbf{1.000} & \textbf{1.000} \\
CharacterTrajectories & 0.990 & 0.960 & 0.994 & 0.990 & \textbf{0.996} & 0.987 & 0.924 & 0.964 & 0.958 & 0.976 & 0.979 & 0.967 & 0.989 & 0.985 \\
Cricket & \textbf{1.000} & 0.986 & 0.986 & 0.986 & 0.986 & 0.986 & 0.931 & 0.986 & \textbf{1.000} & 0.944 & \textbf{1.000} & 0.986 & 0.972 & 0.972 \\
DuckDuckGeese & 0.600 & 0.440 & \textbf{0.720} & 0.680 & 0.640 & 0.480 & 0.380 & 0.460 & 0.460 & 0.480 & 0.440 & 0.580 & 0.360 & 0.340 \\
ERing & 0.930 & 0.926 & 0.981 & 0.985 & 0.919 & 0.919 & 0.856 & 0.981 & 0.978 & 0.952 & \textbf{0.993} & \textbf{0.993} & 0.970 & 0.959 \\
Epilepsy & 0.964 & 0.949 & \textbf{1.000} & \textbf{1.000} & 0.964 & 0.978 & 0.964 & \textbf{1.000} & \textbf{1.000} & 0.978 & \textbf{1.000} & \textbf{1.000} & \textbf{1.000} & \textbf{1.000} \\
EthanolConcentration & 0.323 & 0.293 & 0.471 & 0.548 & 0.262 & 0.327 & 0.274 & 0.380 & 0.300 & 0.278 & 0.414 & 0.414 & \textbf{0.624} & 0.612 \\
FingerMovements & 0.530 & 0.530 & 0.540 & 0.550 & 0.530 & 0.500 & 0.510 & 0.510 & 0.570 & 0.590 & 0.560 & 0.510 & 0.580 & \textbf{0.610} \\
HandMovementDirection & 0.231 & 0.189 & 0.365 & 0.216 & \textbf{0.392} & 0.270 & 0.351 & 0.297 & 0.270 & 0.378 & 0.324 & 0.311 & 0.324 & 0.324 \\
Handwriting & 0.607 & 0.272 & 0.521 & 0.466 & \textbf{0.667} & 0.332 & 0.208 & 0.279 & 0.266 & 0.269 & 0.206 & 0.281 & 0.455 & 0.422 \\
Heartbeat & 0.717 & 0.727 & 0.756 & 0.761 & 0.751 & 0.771 & 0.722 & 0.741 & 0.722 & 0.732 & 0.785 & \textbf{0.829} & 0.732 & 0.717 \\
JapaneseVowels & 0.949 & 0.759 & 0.981 & 0.978 & \textbf{0.992} & 0.946 & 0.641 & 0.805 & 0.819 & 0.786 & 0.957 & 0.954 & 0.824 & 0.784 \\
LSST & 0.551 & 0.607 & 0.638 & 0.607 & 0.194 & 0.639 & 0.548 & 0.604 & 0.579 & 0.552 & 0.565 & \textbf{0.663} & 0.592 & 0.604 \\
Libras & 0.872 & 0.844 & 0.911 & \textbf{0.933} & 0.878 & 0.833 & 0.678 & 0.867 & 0.883 & 0.800 & 0.906 & 0.928 & 0.889 & 0.872 \\
MotorImagery & 0.500 & 0.510 & 0.520 & 0.500 & 0.550 & 0.490 & 0.530 & 0.550 & 0.540 & 0.530 & \textbf{0.580} & 0.490 & 0.480 & 0.570 \\
NATOPS & 0.883 & 0.839 & 0.933 & 0.872 & \textbf{0.961} & 0.944 & 0.750 & 0.822 & 0.850 & 0.850 & 0.867 & 0.878 & 0.811 & 0.856 \\
PEMS-SF & 0.711 & \textbf{1.000} & 0.821 & 0.803 & 0.780 & 0.994 & 0.902 & 0.988 & \textbf{1.000} & \textbf{1.000} & \textbf{1.000} & 0.988 & 0.942 & 0.954 \\
RacketSports & 0.803 & 0.803 & 0.882 & \textbf{0.914} & 0.895 & 0.888 & 0.783 & 0.836 & 0.849 & 0.796 & 0.908 & \textbf{0.914} & 0.882 & 0.895 \\
SelfRegulationSCP1 & 0.775 & 0.727 & \textbf{0.908} & 0.867 & 0.829 & 0.778 & 0.720 & 0.788 & 0.812 & 0.795 & 0.788 & 0.819 & 0.877 & 0.887 \\
SelfRegulationSCP2 & 0.539 & 0.550 & 0.533 & \textbf{0.589} & 0.483 & 0.522 & 0.511 & 0.522 & 0.522 & 0.439 & 0.522 & 0.517 & 0.439 & 0.439 \\
StandWalkJump & 0.200 & 0.400 & 0.400 & 0.400 & 0.400 & 0.133 & 0.400 & \textbf{0.667} & 0.467 & 0.600 & 0.467 & 0.467 & 0.400 & 0.400 \\
UWaveGestureLibrary & 0.903 & 0.872 & \textbf{0.931} & 0.909 & 0.903 & 0.887 & 0.759 & 0.900 & 0.881 & 0.912 & 0.906 & 0.900 & 0.881 & 0.866 \\
\midrule
\textit{Average} & 0.698 & 0.680 & \textbf{0.747} & 0.737 & 0.715 & 0.700 & 0.649 & 0.720 & 0.708 & 0.702 & 0.726 & 0.733 & 0.714 & 0.724 \\
\textit{Mean rank} & 8.65 & 10.10 & \textbf{4.79} & 5.48 & 6.96 & 8.04 & 11.54 & 7.21 & 7.40 & 8.29 & 5.31 & 5.83 & 7.94 & 7.46 \\
\bottomrule
\end{tabular}%
}
\end{table}

\end{document}